\pdfoutput=1

\documentclass[11pt]{article}

\usepackage{acl}

\usepackage{times}
\usepackage{latexsym}
\usepackage{amsmath,amssymb}
\usepackage{graphicx}
\usepackage{geometry}
\usepackage{booktabs, multirow}
\usepackage{devanagari}
\usepackage{amsmath}
\usepackage{algorithm}
\usepackage{algpseudocode}
\usepackage[none]{hyphenat}

\usepackage[T1]{fontenc}

\usepackage[utf8]{inputenc}

\usepackage{microtype}

\title{\textsc{CharSpan:} Utilizing Lexical Similarity to Enable Zero-Shot Machine Translation for Extremely Low-resource Languages}

\author{Kaushal Kumar Maurya\textsuperscript{1,3}\thanks{\; Work done during  first author's internship at Microsoft. He was enrolled as a graduate student at IIT Hyderabad at that time.} \and
    Rahul Kejriwal\textsuperscript{2}  \\
    \textbf{Maunendra Sankar Desarkar\textsuperscript{1}} \and 
    \textbf{Anoop Kunchukuttan\textsuperscript{2}} \\
    \textsuperscript{1}NLIP Lab, IIT Hyderabad, India \\
    \textsuperscript{2}Microsoft, India    \textsuperscript{3}MBZUAI, UAE\\
    \href{mailto:cs18resch11003@iith.ac.in}{cs18resch11003@iith.ac.in}, \href{mailto:maunendra@cse.iith.ac.in}{maunendra@cse.iith.ac.in}\\ \href{mailto:\{rahul.kejriwal,anoop.kunchukuttan\}@microsoft.com}{\{rahul.kejriwal, anoop.kunchukuttan\}@microsoft.com} \\
}

\begin{document}
\maketitle
\begin{abstract}
We address the task of machine translation (MT)
from extremely low-resource language (ELRL)
to English by leveraging cross-lingual transfer
from \textit{closely-related} high-resource language (HRL). The development of an MT system for ELRL is challenging because these languages typically lack parallel corpora and monolingual corpora, and their representations are absent from large multilingual language models. Many ELRLs share lexical similarities with some HRLs, which presents a novel modeling opportunity. However, existing subword-based neural MT models do not explicitly harness this lexical similarity, as they only implicitly align HRL and ELRL latent embedding space. To overcome this limitation, we propose a novel, \textsc{CharSpan}, approach based on \textit{character-span noise augmentation} into the training data of HRL. This serves as a regularization technique, making the model more robust to \textit{lexical divergences} between the HRL and ELRL, thus facilitating effective cross-lingual transfer. Our method significantly outperformed strong baselines in zero-shot settings on closely related HRL and ELRL pairs from three diverse language families, emerging as the state-of-the-art model for ELRLs.

\end{abstract}

\section{Introduction}
\label{sec:intro}


Recent advancements in multilingual modeling have expanded the coverage of Natural Language Processing (NLP) technologies to many LRLs by transferring knowledge from HRLs to LRLs. As a result, this progress has led to remarkable advancement in multiple NLP tasks, including MT, transliteration, natural language understanding, and text generation \citep{johnson-etal-2017-googles,kunchukuttan-etal-2018-leveraging,conneau-etal-2020-unsupervised,liu-etal-2020-multilingual-denoising} for LRLs. However, most of the existing work has focused on the top few hundred languages represented on the web \cite{joshi-etal-2020-state}. The availability of monolingual corpora and/or parallel corpora for these languages has been the driving force behind this progress, achieved either through direct training, few-shot training, or learning with large multilingual language models (mLLMs). This enables learning common embedding spaces that facilitate cross-lingual transfer \cite{nguyen-chiang-2017-transfer,khemchandani-etal-2021-exploiting}. However, there is a long tail of languages for which no monolingual or parallel corpora are available, and they are absent from mLLMs. These languages are referred to as ELRLs. This paper is a step toward building MT systems for ELRLs.

\begin{figure}
    \captionsetup{font=scriptsize}
    \centering
     \scriptsize
    \includegraphics[scale=0.50]{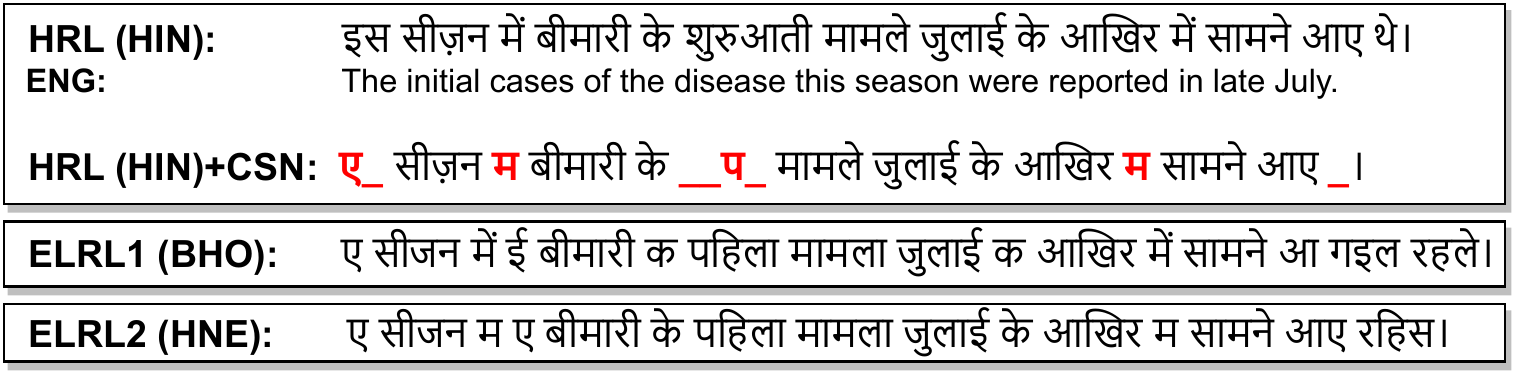}
    \vspace*{-0.4cm}
    \caption{Hindi (HIN; HRL), Bhojpuri (BHO; ELRL) and Chhattisgarhi (HNE; ELRL) parallel sentences. Additionally, the corresponding noisy Hindi example with character-span noise. BHO and HNE are closely related to HIN.}
    \label{fig:sample_poster}
\vspace*{-0.6cm}
\end{figure}

Fortunately, many of ELRLs are lexically similar to some HRLs. \textit{Lexical similarity refers to languages sharing words with similar form (spelling and pronunciation) and
meaning}.\footnote{\url{https://en.wikipedia.org/wiki/Lexical\_similarity}} This includes cognates, lateral borrowings and loan
words. We explore if cross-lingual transfer can be enabled or improved for ELRLs by \textit{explicitly} taking lexical similarity into account. In particular, \textit{we explore MT from an ELRL to another language (English) with transfer enabled by a related HRL on the source side.} Our key \textit{insight} is that cognates in ELRL having similar spelling to the HRL word can be thought of as misspellings of the latter. For example, the word {\dn lgtA} (\textit{lagta}) in Hindi (HRL) is spelled as {\dn lAgatA} (\textit{laagata}) in Bhojpuri (LRL). If we make the HRL model robust to spelling variations, it will improve cross-lingual transfer to related ELRLs. To achieve spelling variation robustness, we propose novel \textit{character-span noise augmentation (\textsc{CharSpan})} in the HRLs training data. A sample example is presented in Fig. \ref{fig:sample_poster}. This acts as a regularizer and makes the model more robust to perturbations in representations of words in closely related languages and improves model generalization for lexically similar languages.


Our key contributions are: (1) We propose a novel model \textsc{CharSpan:} \textit{\underline{Char}acter-\underline{Span} noise augmentation}, which considers surface level lexical similarity to improve cross-lingual transfer between closely-related HRLs and LRLs. The proposed approach shows a 12.5\% chrF improvement over baseline NMT models across all considered ELRLs. Our model also shows performance improvement over various data augmentation baselines. (2) We show that our approach generalizes across three typologically diverse language families, comprising 6 HRLs and 12 ELRLs. (3) We provide detailed ablation and analysis to gain insights and demonstrate the effectiveness of our approach.

\begin{figure}
    \captionsetup{font=scriptsize}
    \centering
     \scriptsize
    \includegraphics[scale=0.65]{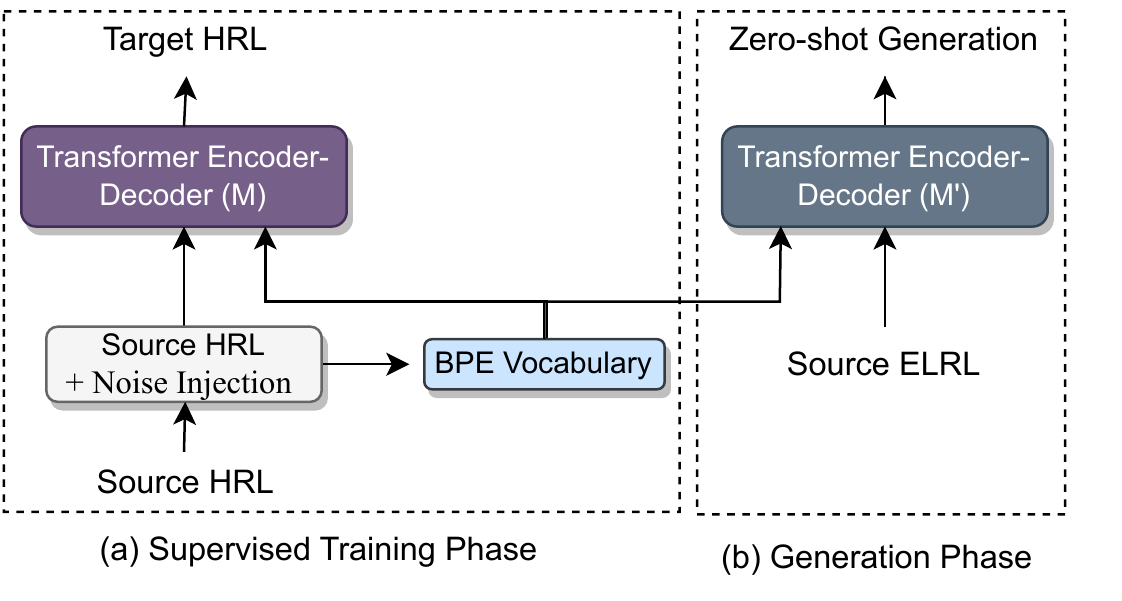}
    \vspace{-0.25cm}
    \caption{Overview of proposed \textsc{CharSpan} model}
    \label{fig:model_arch}
    \vspace{-0.5cm}
\end{figure}

\section{Related Work}
\label{sec:related}


Traditionally, character-level noise has been used to improve the robustness of MT systems to spelling mistakes and ASR errors \citep{sperber2017toward,vaibhav-etal-2019-improving,karpukhin-etal-2019-training}. However, these approaches are mostly investigated for their impact on robustness rather than for cross-lingual transfer. More recently, token/BPE-level general noise augmentation approaches such as WordDropout \cite{sennrich-etal-2016-edinburgh} and  SwitchOut \cite{wang-etal-2018-switchout} have been proposed, but they have limited cross-lingual transfer capabilities.  Close to our work, \citet{aepli-sennrich-2022-improving} and \citet{blaschke-etal-2023-manipulating} show that augmenting data with character-level noise can help cross-lingual transfer. The models were evaluated with NLU tasks. n contrast, our work focuses on MT, an NLG task, which is much more challenging than an NLU task in a zero-shot setting. Furthermore, we explore span noise augmentation, which considers larger lexical divergence (less lexical similarity between the HRL and ELRL) and enables better cross-lingual transfer. 

In other work on utilizing lexical similarity,  \citet{patil-etal-2022-overlap} proposed OverlapBPE, which takes lexical overlap between HRL and LRL into account while learning BPE vocabulary. \citet{provilkov-etal-2020-bpe} introduced BPE-Dropout, providing on-the-fly non-deterministic segmentations while training. Soft Decoupled Encoding (SDE) \citet{DBLP:conf/iclr/WangPAN19} utilizes lexical information without pre-segmenting the data by decoupling the lexical and semantic representations. SDE requires small monolingual data for modeling. In contrast, the \textsc{CharSpan} model does not require any training resources for ELRLs. It only needs script similarity between the HRL and ELRL.



\section{The \textsc{CharSpan} Model}

Figure \ref{fig:model_arch} presents an overview of the proposed \textsc{CharSpan} model, for ELRL to English MT task. The model has two phases: supervised training with noisy HRL and zero-shot generation with ELRLs.
\\
\noindent\textbf{Model Training and Generation:} In the \textit{supervised training phase}, the source-side training data of the HRL pair ($\mathcal{D_H}$) is augmented with character-span noise (described later) to create the augmented parallel corpus ($\mathcal{D^{'}_H} = \eta(\mathcal{D_H})$), where $\eta$ is the noise function. $\eta(\mathcal{D_H})$ can be considered as the proxy parallel data for the ELRL-English translation task. Next, we learn a subword vocabulary ($\mathcal{V}$) using $\mathcal{D^{'}_H}$, i.e., the noise is augmented before learning the vocabulary. A standard encoder-decoder transformer model ($\mathcal{M}$; \citet{vaswani2017attention}) is then trained with $\mathcal{D^{'}_H}$ and $\mathcal{V}$ from scratch in a supervised setting to obtain the trained model $\mathcal{M^{'}}$. Finally, in the \textit{zero-shot generation phase}, for a given source ELR language $\mathcal{L}$, the target English translation is obtained using $\mathcal{M^{'}}$ and $\mathcal{V}$ in the zero-shot setting.
\\
\noindent \textbf{Character Span Noise Function:} The noise functions serve to make the model robust to spelling variations between related languages. This acts as a regularizer and helps improve cross-lingual representation and transfer. Intuitively, the existing unigram character noise might address limited lexical variations between HRL and ELRLs. \textit{To address larger lexical divergence, we propose a \textsc{CharSpan} where span noise is augmented}. Formally, for a given sentence, $x \in \mathcal{X}$ from $\mathcal{D_H(X,Y)}$ with indices $I = {1, 2, \ldots, |x|}$, a subset of these indices $I_s \subset I$ is randomly and uniformly selected as the starting point for the noise augmentation. Subsequently, 1-3 character gram spans are iteratively sampled until the noise augmentation budget (i.e., 9\% - 11\% characters) is exhausted. We employ \textit{span deletion} and \textit{span replacement with a single random character of ELRL}, both with equal probability as the noising operations\footnote{We explored some linguistically motivated noising schemes, but these were not beneficial.}. This \textsc{CharSpan} is inspired by SpanBERT \cite{joshi2020spanbert}\footnote{SpanBERT applies denoising to subword tokens while we apply it at the character level.}. A formal algorithm is presented in the Algorithm \ref{algo:noise_injection}. We conducted experiments with all three operations (including insertion), with different percentages of noise and various other experimental setups, as outlined in Appendix Table \ref{tab:ablation}. We found the presented noise augmentation configuration to be the most effective.

\begin{algorithm*}
\small
\caption{\textsc{CharSpan:} Character-span Noise Augmentation Algorithm}
\begin{algorithmic}[1]
\Require \textbf{[Inputs]} high resource language data ($\mathcal{D_H (X, Y)}$) from $H$-$En$ parallel corpus, range of noise augmentation percentage $[P1,  P2]$, set of noise augmentation candidates $C$ (see Fig. \ref{tab:candalpha}), largest character $n$-gram size $N$ that will be considered for noising
\Ensure \textbf{[Output]} Noisy high resource language data ($\mathcal{D^{'}_H}$)
\State Augmentation percentage ($I_p$) $=$  random float(P1, P2) \# find a random float value between $P1$ and $P2$
\State Augmentation factor ($\alpha$) $=$ int($I_p / N$)
\For{each $h$  in $\mathcal{X}$ }
    \State Let $sz$ be the number of characters in $h$.
    \State Let $Indices=\{\lceil(N/2)\rceil, \cdots, sz-\lceil(N/2)\rceil\}$ \# Leaving $\lceil(N/2)\rceil$ character indices from beginning and end
    \State Randomly select $S = N * \alpha$ character indices from $Indices$
    \For{each $k$ in $S$ }
        \State Span gram ($Sp_N$) $=$ sample character-span size uniformly from $\{1, 2, \dots, N\}$ with equal probability
        \State Operation  $(O_p)$ $=$ sample operations uniformly from $\{$ delete, replace $\}$ with equal probability
        \State $C_d$ =\{\}
        \If{$(O_p)$ is replace}
            \State Candidate char ($c$) $=$ single sample character uniformly from $C$ with equal probability
            \State Append candidate char $c$ in $C_d$
        \EndIf
        \If{$Sp_N ==$ 1}
            \State Perform the operation $(O_p)$ with $C_d$ at the index $k$ 
        \Else  
             \State Perform the operation  $(O_p)$   with $C_d$ at the indexes from $k - int((Sp_N - 1) / 2)$ to $ k + int((Sp_N - 1) / 2)$
        \EndIf
        \EndFor
    \EndFor
        
\end{algorithmic}
\label{algo:noise_injection}
\end{algorithm*}


\section{Experimental Setup}

We seek answers to the following questions:  (1) Does the span noise augmentation improve cross-lingual transfer, i.e., zero-shot performance for related ELRLs for MT task? (2) Why does the model's cross-lingual transfer improve? - Insights from the learned embedding space. (3) Is the proposed approach scalable to typologically diverse language families? 

\subsection{Datasets and Languages}
We evaluated the performance of the proposed model on three language families: Indo-Aryan, Romance, and Malay-Polynesian. We considered six HRLs and twelve LRLs (two HRLs and several ELRLs from each family). All the ELRLs are lexically similar and have the same script with corresponding HRLs, as shown in Figure \ref{fig:langsim} (Appendix \ref{app:langsim}). Parallel training data for the HRLs was selected from publicly available datasets. The model's performance was evaluated on the FLORES-200 devtest set \cite{costa2022no}. Dataset statistics are presented in the Appendix.

\subsection{Baselines and Evaluation Metrics}

Based on recent literature in low-resource MT, we compare our approach with the following strong baselines: (a) Vanilla NMT with BPE segmentation (BPE; \citet{sennrich-etal-2016-neural}), (b) General data augmentation methods: (Sub)WordDropout and (Sub)WordSwitchOut, (c) Methods using lexical similarity: Overlap BPE, BPE-Dropout, SDE and unigram char-noising \cite{aepli-sennrich-2022-improving}. Baselines and model training details are provided in Appendix. Following recent studies on MT for ELRLs \cite{costa2022no, siddhant2022towards}, we use chrF \cite{popovic-2015-chrf} as the primary evaluation metric. In addition, we also report BLEU \cite{papineni-etal-2002-bleu} and two neural metrics viz., BLEURT \cite{sellam-etal-2020-bleurt} and COMET \cite{rei-etal-2020-comet} scores in Appendix \ref{sec:othermetirc}. 

\section{Results and Analyses}

\begin{table*}[ht]
\centering
\resizebox{0.9\textwidth}{!}{  
\begin{tabular}{|l| cccccccc | cc |cc | c|}
    \hline
\multirow{2}{*}{\textbf{Models}} 
        & \multicolumn{8}{c|}{\textbf{ Indo-Aryan}} &
        \multicolumn{2}{c|}{\textbf{Romance}} & \multicolumn{2}{c|}{\textbf{Malay-Polynesian}} & \multirow{2}{*}{\textbf{Average}}  \\
        \cline{2-13}
        & \textbf{Gom}  & \textbf{Bho}  & \textbf{Hne}  & \textbf{San}  & \textbf{Npi}  & \textbf{Mai}  & \textbf{Mag} &  \textbf{Awa} & \textbf{Cat} & \textbf{Glg} & \textbf{Jav} & \textbf{Sun} &          \\
    \hline
    BPE* & 26.75 & 39.75 & 46.57 & 27.97 & 30.84 & 39.79 & 48.08 & 46.28 & 33.32 & 53.75 & 31.44 & 32.21 & 38.06     \\
    \hline
    WordDropout &  27.01 & 39.57 & 46.19 & 28.13 & 31.91 & 40.31 & 47.37 & 46.48 & 34.20 & 52.21 & 32.03 & 32.52 & 38.16
  \\
    SubwordDropout &  27.91 & 40.11 & 46.26 & 29.46 & 32.56 & 40.99 & 47.91 & 47.43 & 35.09 & 52.28 & 33.38 & 33.47 & 38.90
  \\
    WordSwitchOut &  25.17 & 38.81 & 45.87 & 26.21 & 29.95 & 39.69 & 47.53 & 44.54 & 32.98 & 51.81 & 31.84 & 32.49 & 37.24
 \\
    SubwordSwitchOut & 26.08 & 38.84 & 45.84 & 28.19 & 30.81 & 40.19 & 47.28 & 45.93 & 33.26 & 53.71 & 31.24 & 32.06 & 37.78  \\
    \hline
    OBPE   & 27.90 & 40.57 & 47.46 & 28.52 & 31.99 & 40.71 & 49.10 &  47.16 & 32.33 & 52.77 & 29.98 & 30.88 & 38.28  \\
    SDE & 28.01 & 40.91 & 47.88 & 28.66 &  32.03  & 40.82 & 48.96 &  47.30  & 33.72  & 53.95 & 31.84 & 31.24 & 38.77     \\

    BPE-Dropout*   & 28.65 & 40.84 & 46.58 & 28.80 & 31.88 & 40.79 & 47.86 & 47.32  & 34.56 & 55.83 & 32.01 & 32.97 & 39.00         \\
    \hline
    unigram char-noise** & 28.85 & 42.53 & 49.35 & 29.80 & 34.61 & 42.67 & 50.97 & 49.43 & 43.16 & 54.81 & 35.42 & 36.69 & 41.52         \\
    BPE $\rightarrow$ SpanNoise*** $(ours)$ & 28.66 & 41.94 & 49.48 & 30.49 & 35.66 & 44.75 & 50.55 & 49.21  & 43.11 & 54.89 & 36.12  & 37.11  &  40.16        \\  
    \textsc{CharSpan} $(ours)$ & 29.71 & 43.75 & 51.69 & \underline{\textbf{31.40}} & 36.52 & 45.84 & 51.90 & 50.55  & 43.51 & 55.46 & 36.24 & 37.31 &  42.82  \\
    \textsc{CharSpan} + BPE-Dropout $(ours)$& \underline{\textbf{29.91}} & \underline{\textbf{44.02}} & \underline{\textbf{51.86}} & 30.88 & \underline{\textbf{37.15}} & \underline{\textbf{46.52}} & \underline{\textbf{52.99 }} & \underline{\textbf{51.34}} & \underline{\textbf{44.93}} & \underline{\textbf{ 55.87}} & \underline{\textbf{ 36.97}} & \underline{\textbf{ 38.09}} & \underline{\textbf{ 43.37}}    \\  
\hline
\end{tabular}}
\caption{\small Zero-shot chrF scores results for ELRLs $\rightarrow$ English machine translation. We conducted statistical significance tests to compare \textsc{CharSpan} with the diverse baselines: BPE, BPE-Dropout, Unigram char-noise, and BPE $\rightarrow$ SpanNoise, using paired bootstrap sampling \cite{post-2018-call}. \textsc{CharSpan} improvements over these baselines are statistically significant with *($p < 0.0001$), **($p < 0.001$), and *** ($p < 0.05$). Similar observations hold across other evaluation metrics presented in the Appendix.}
\label{tab:resultschrf}
\end{table*}

The proposed \textsc{CharSpan} and baseline models' results across different language families are presented in Table \ref{tab:resultschrf}. The following are the major observations:\\


\noindent\textbf{Noise vs. Baselines:} All the proposed noise augmentation models outperform vanilla NMT and all baseline models that utilize lexical similarity (i.e., OBPE, BPE-Dropout, and SDE). This trend is consistent across all language families and ELRLs. Moreover, existing lexical similarity-based baselines do not provide any major improvement in translation quality over vanilla NMT. Possible reasons for this can be twofold: (1) most of the ELRLs either do not have monolingual data (OBPE and SDE are required) or have small data, and (2) we observe that in OBPE, approximately 90\% of vocabulary tokens are already overlapping among HRLs and ELRLs, leaving little room for learning additional overlapping tokens. This is expected, as these two language sets are closely related. The proposed \textsc{CharSpan} method also outperforms general data augmentation methods like (Sub)WordDropout and (Sub)WordSwitchout, showing its effectiveness.\\

\noindent\textbf{Unigram vs. Char-Span Noise:} We are first to explore unigram char noise \cite{aepli-sennrich-2022-improving} for related language MT. We see that unigram char noise is beneficial for the task. However, our proposed \textsc{CharSpan} provides significant improvements over unigram character noise. We believe our proposed data augmentation is more effective in bringing language representations closer. \\ 

\noindent\textbf{When to introduce noise?} To understand when noise augmentation is effective, we augmented noise after learning the vocabulary in the baseline (BPE $\rightarrow$ SpanNoise). This leads to improved performance over all baselines. This enables scalability since augmenting noise after learning the vocabulary allows the application of this method to large language models that have fixed vocabulary. However, the results suggest that applying noise prior to learning the vocabulary, as in \textsc{CharSpan}, yields slightly better results. Further, we conducted statistical significance tests to compare BPE $\rightarrow$ SpanNoise with BPE, BPE-Dropout, and Unigram char-noise baselines using paired bootstrap sampling \cite{post-2018-call}. We observed that the BPE $\rightarrow$ SpanNoise model is superior to the baseline BPE and BPE-Dropout methods (statistically significant at $p < 0.001$), demonstrating that adding noise after segmentation is also highly effective. Additionally, we noticed that BPE $\rightarrow$ SpanNoise outperforms unigram char-noise for 11 out of 12 languages at $p < 0.05$. Thus, introducing character span noise after segmentation provides a statistically significant improvement over baselines, which can be advantageous when working with pre-trained models.\\

\noindent\textbf{Combining noise and BPE-dropout:} We see that combining \textsc{CharSpan} with BPE-dropout gives the best-performing results. \\

    

\begin{table}
\centering
\resizebox{0.45\textwidth}{!}{  
\begin{tabular}{|l| c| c| c|c|}
    \hline
     \textbf{Langs.} &  \textbf{BPE}  &  \textbf{Unigram Noise}  & \textbf{Char-Span Noise}      & \textbf{Sim }\\
    \hline
    Guj-Deva &  34.36 &  36.17 &  38.09  &  0.42    \\
    Pan-Deva  &  29.18 &  33.34 &  36.50   &  0.40   \\
    Ben-Deva &  25.35 &  28.42 &  30.28  &  0.34  \\
    Tel-Deva &  23.30 &  24.05 &  24.12   & 0.27   \\
    Tam-Deva &  13.81 &  13.69 &  14.40  & 0.15    \\
    
\hline
\end{tabular}}
\caption{\small Zero-shot chrF scores with additional lexically less similar languages. \texttt{HRL:} hi and mr; \texttt{sim:} lexical similarity}
\label{tab:analysis1}
\vspace{-0.5cm}
\end{table}

\noindent\textbf{Performance on Less Similar Languages:} We evaluate the model's performance on languages that are less lexically similar to the considered languages and have different scripts. The languages are Gujarati (Guj), Punjabi (Pan), Bengali (Ben), Telugu (Tel), and Tamil (Tam). We first perform script-conversion of these languages to HRL by \citet{kunchukuttan2020indicnlp}). The training setup is similar to the Indo-Aryan family. Table \ref{tab:analysis1} shows that the ELRLs, which are lexically similar to HRLs, demonstrate a larger performance gain, while those with less lexical similarity show limited improvement. This suggests that the model's effectiveness is closely tied to the lexical similarity of the languages in \textsc{CharSpan}. \\


\noindent\textbf{Impact of Cross-lingual Transfer:} In this analysis, we investigate the encoded representations of the sentences to gain insights into how performance improves with char-span noise augmentation. We collected pooled last-layer representations of the encoder for HRL and LRLs across all parallel test examples using BPE, unigram char-noise (UCN), and the \textit{CharSpan} models. We then calculated the average cosine similarity scores across the test set, presented in Table \ref{tab:clt}. Notably, the \textit{CharSpan} model demonstrates high similarity, indicating a well-aligned embedding space for enhanced cross-lingual transfer.\\

\begin{table}[h]
\centering
\resizebox{0.45\textwidth}{!}{  
\begin{tabular}{|l| ccccccc | }
    \hline
        \textbf{Models}  & \textbf{Bho}  & \textbf{Hne}  & \textbf{San}  & \textbf{Npi}  & \textbf{Mai}  & \textbf{Mag} &  \textbf{Awa}        \\
    \hline
    BPE & 0.761 & 0.793 & 0.701 & 0.744 & 0.762 & 0.809 & 0.792   \\
     UCN & 0.853 & 0.888 & 0.765 & 0.821 & 0.849 & 0.897 & 0.883   \\
   \textsc{CharSpan}  & \textbf{0.871} &\textbf{0.909} & \textbf{0.789} & \textbf{0.858} & \textbf{0.868} & \textbf{0.913} & \textbf{0.901}  \\
\hline
\end{tabular}}
\caption{\small Average cosine similarity between representations of source HRLs and source ELRLs for Indo-Aryan family. Results for other families are in the Appendix \ref{sec:cltfull}.}
\label{tab:clt}
\vspace{-0.3cm}
\end{table}

\noindent\textbf{Importance of Selecting Right HRLs:} Table \ref{tab:analysis2} presents an analysis of the impact of lexically diverse  HRLs used for training. Results indicate that the \textsc{CharSpan} model demonstrates a performance gain when lexically similar HRLs were considered for noise injection. When the HRLs are less lexically similar, a degradation in performance is observed. These findings indicate the importance of using lexically similar HRLs.     \\


\begin{table}[!htb]
\centering
\scriptsize
\begin{tabular}{|l|c|c|c|c|c|}
    \hline
    \textbf{Model}    &   \textbf{Hne}  & \textbf{Mag} &  \textbf{Mai} & \textbf{Npi} & \textbf{San}   \\
    \hline
    \multicolumn{6}{|c|}{\textit{Training with Lexically Similar HRLs: Hin, Mar, Pan, Guj, Ben}} \\
    \hline 
    BPE &  43.04 &   45.08&  39.51  & 31.92 & 29.29  \\
    Char-span Noise &  45.89 &  45.82 &   41.67 & 34.40 & 30.34  \\
    \hline 
    \multicolumn{6}{|c|}{\textit{Training with Lexically less similar HRLs: Hin, Tel, Tam, Mal, Ora}} \\
    \hline
    BPE &  41.87 &    42.27&   36.95 & 30.50 & 26.95\\
    Char-span Noise &  39.93 &   40.34 &  37.98 & 29.20 & 25.84\\
\hline
\end{tabular}
\caption{\small Analysis experiment to show zero-shot chrF scores with lexically diverse HRLs. Due to computational constraints, we have considered 1 million parallel data for each HRL.}
\label{tab:analysis2}
\end{table}

\noindent\textbf{Impact of small ELRL parallel Data:} Here, we combined small ELRLs parallel data with the HRLs training data for BPE and \textsc{CharSpan} model. The results are presented in Table \ref{tab:addparelrls} in the appendix \ref{sec:smallpd}. The additional data boosts both model performance, and \textsc{CharSpan} still outperforms the BPE model.\\
\\
\noindent\textbf{Error Analyses:} In Appendix \ref{sec:errana}, we have conducted two error analyses: \textit{Transliteration Errors} and \textit{Grammatical Well-formedness}. In Fig.~\ref{fig:erroranalysis1}, it can be observed that the unigram model often performs transliteration instead of translation for many input words. However, the proposed model does not encounter such errors, and the impact of transliteration errors is minor. This observation holds across test data. This is possible because \textsc{CharSpan} augments the span, resulting in stronger regularization and enabling more contextual zero-shot cross-lingual transfer. In Table \ref{tab:errorana2}, there is a comparison of sentence well-formedness, indicating that zero-shot generations for the unigram model, as opposed to CharSpan, are not grammatically well-formed. \\

\vspace{-0.5cm}
\section{Conclusion}

This study presents a simple yet effective novel character-span noise argumentation model, \textsc{CharSpan}, to facilitate better cross-lingual transfer from HRLs to closely related ELRLs. The approach generalizes to closely related HRL-ELRL pairs from three typologically diverse language families. The proposed model consistently outperformed all the baselines. To the best of our knowledge, we are the first to apply noise augmentation for the NLG task. In the future, we will extend \textsc{CharSpan} to other NLP tasks, combine it with pre-trained models, and investigate noise augmentation in English-to-ELRL MT task.


\section*{Limitations}
The current work only addresses cross-lingual transfer during translation from ELRLs to English. It still remains to be investigated if noise augmentation is beneficial for translation from English to extremely low-resource languages. We assume that the related languages also use the same script or scripts that can be easily mapped/transliterated to each other. This method might not be effective for transfer between related languages that are written in very different scripts e.g. Hindi is written in the Devanagari script, while Sindhi is written in the Perso-Arabic script.

\section*{Ethics Statement}

We have formulated low-resource languages as a misspelled version of a high-resource language. We would like to clarify that our suggestion is not that the low-resource languages are misspelled versions of higher-resource-related languages. This is not a \textit{linguistic claim}, and as would be evident from comparative linguistics, most such scenarios are likely co-evolutions of related languages. This perspective of related languages is only a \textit{technical tool} to make use of the fact that the end result of the co-evolution of related languages is that they ``\textit{look like}'' spelling variations of each other, and hence, robustness methods applied to NMT can be adapted for this scenario.

This work did not involve any new data collection and did not employ any annotators for data collection. We use publicly available datasets for experiments reported in this work. Some of these datasets originate from webcrawls and we do not make any explicit attempt to identify any biases in these datasets and use them as-is.

\bibliography{custom}

\appendix

\section{Baselines}
\label{sec:baselineapp}
We compare the proposed model performance with the following strong baselines:

\begin{itemize}
    \item \textbf{Vanilla NMT (BPE; \citet{sennrich-etal-2016-neural}):} Neural Machine Translation model training with the standard BPE algorithm.
    \item \textbf{WordDropout  \cite{sennrich-etal-2016-edinburgh}:} In this baseline, randomly selected words in the source/target sentence have their embeddings set to 0. We have selected 10\% words in the source sentence as the noise augmentations are done in the source.   
    \item \textbf{SubwordDropout:} It is a variant of WordDropout baseline where we drop the BPE tokens instead of words.
    \item \textbf{WordSwitchOut \cite{wang-etal-2018-switchout}:} This baseline employs a data augmentation technique where random words in both the source and target sentences are replaced with randomly selected words from their respective vocabularies. We have utilized the officially released implementation with a 10\% word replacement rate. 
    \item \textbf{SubwordSwitchOut:} It is a variant of WordSwitchOut baseline where we use the BPE tokens instead of words.
    \item \textbf{Overlap BPE (OBPE; \citet{patil-etal-2022-overlap}):} The approach modifies the BPE algorithm to encourage more shared tokens between high-resource and low-resource languages tokens in the vocabulary. This model required a monolingual dataset for ELRLs. We use a small monolingual dataset, based on availability, for the ELRLs. Earlier work applied OBPE for NLU tasks only - we are the first to investigate it for MT. 
    \item \textbf{Soft Decoupled Encoding (SDE; \cite{DBLP:conf/iclr/WangPAN19}):} In the SDE approach, the authors have designed a framework that effectively decouples word embeddings into two interacting components: representing the spelling of words and capturing the latent meaning of words. This modeling technique has demonstrated its effectiveness in improving the performance of low-resource languages. In our study, we utilized the officially released implementation of SDE.  
    \item \textbf{BPE-Dropout \cite{provilkov-etal-2020-bpe}:} It utilizes the BPE algorithm to learn the vocabulary and sample different segmentations for input text during training (on-the-fly).
    \item \textbf{Unigram Character Noise (UCN; \citet{aepli-sennrich-2022-improving}):} Inspired by the UCN model, we augment character-level noise (with all three operations) instead of char-span, the rest of the setup is similar to \textsc{CharSpan}.
    \item \textbf{BPE $\rightarrow$ Char-Span Noise:} In this ablation,  we first learn vocabulary with clean HRLs. Subsequently, character-span noise is augmented into training data. This will demonstrate the significance of learning the BPE vocab with the noisy dataset.
    \item \textbf{Char-Span Noise + BPE-Dropout:} In this model, we train the BPE-Dropout model with char-span noise augmented HRLs training dataset.
    
\end{itemize}

\begin{table*}
\scriptsize
\centering
\scalebox{0.97}{
\begin{tabular}{ |l|l|l|l|l|l|l|c|c|c| l| } 
 \hline
 \textbf{Family } & \textbf{Code} &\textbf{ Language} & \textbf{Script} & \textbf{Family} & \textbf{Subgrouping} & \textbf{Res.} & \textbf{Train} & \textbf{Dev }& \textbf{Test} & \textbf{Data Source}\\
 \hline
 \multirow{10}{*}{\textbf{1}}
 & Hin & Hindi & Devanagari & Indo-European & Indo-Aryan & High & 10M & 1000 & 2390 & \citet{ramesh-etal-2022-samanantar} \\ 
 & Mar & Marathi & Devanagari & Indo-European & Indo-Aryan  & High & 3.6M & 1000 & 2390 & \citet{ramesh-etal-2022-samanantar}\\ 
 & Bho & Bhojpuri & Devanagari & Indo-European & Indo-Aryan  & Low & - & - & 1012 & FLORES-200\\ 
& Gom & Konkani & Devanagari & Indo-European & Indo-Aryan  & Low & - & - & 2000 & ILCI\footnote{\url{http://sanskrit.jnu.ac.in/projects/ilci.jsp?proj=ilci}} \\ 
& Hne & Chhattisgarhi & Devanagari & Indo-European & Indo-Aryan  & Low & - & - & 1012 & FLORES-200\\ 
& San & Sanskrit & Devanagari & Indo-European & Indo-Aryan  & Low & - & - & 1012 & FLORES-200 \\ 
& Npi & Nepali & Devanagari & Indo-European & Indo-Aryan  & Low & - & - & 1012 & FLORES-200 \\ 
& Mai & Maithili & Devanagari & Indo-European & Indo-Aryan  & Low & - & - & 1012 & FLORES-200 \\ 
& Mag & Magahi & Devanagari & Indo-European & Indo-Aryan  & Low & - & - & 1012 & FLORES-200 \\ 
& Awa & Awadhi & Devanagari  & Indo-European & Indo-Aryan  & Low & - & - & 1012 & FLORES-200 \\ 
\hline
 \multirow{4}{*}{\textbf{2}}
 & Spa & Spanish & Latin & Indo-European & Romance & High & 6.6M & 670 & 1131 & \citet{rapp-2021-similar} \\ 
 & Pot & Portuguese &Latin & Indo-European & Romance & High & 4.8M & 681 & 1103 & \citet{rapp-2021-similar} \\ 
 & Cat & Catalan & Latin & Indo-European & Romance &  Low & - & - & 1012 & FLORES-200\\ 
 & Glg & Galician & Latin & Indo-European & Romance &  Low  &- & - & 1012 & FLORES-200\\ 
 \hline
\multirow{4}{*}{\textbf{3}}
 & Ind & Indonesian & Latin & Austronesian & Malay-Polynesian & High & 0.5M & 2500 & 3000  & OPUS\footnote{\url{https://opus.nlpl.eu/}} \\ 
 & Zsm & Malay & Latin & Austronesian & Malay-Polynesian & High  & 0.3M & 1500 & 2000 & OPUS\\ 
& Jav & Javanese & Latin & Austronesian & Malay-Polynesian & Low &- & - & 1012 & FLORES-200 \\ 
 & Sun & Sundanese &Latin & Austronesian & Malay-Polynesian & High  & - & - & 1012 & FLORES-200\\ 
 \hline
 \multirow{4}{*}{\textbf{Others}}
 & Pan & Panjabi & Gurmukhi & Indo-European & Indo-Aryan & Low & 1M* & 1000* & 1012  & FLORES-200 \\ 
 & Guj & Gujarati & Gujarati & Indo-European & Indo-Aryan & Low  & 1M* & 1000* & 1012 & FLORES-200 \\ 
& Ben & Bengali & Bengali & Indo-European & Indo-Aryan & High & 1M* & 1000* & 1012 & FLORES-200 \\ 
& Tam & Tamil & Tamil & Indo-European & Indo-Aryan & Low  & 1M* & 1000* & 1012 & FLORES-200\\ 
& Tel & Telugu & Dravidian & Indo-European & Indo-Aryan & Low  & 1M* & 1000* & 1012 & FLORES-200\\ 
& Mal & Malayalam & Malayalam  & Indo-European & Indo-Aryan & Low  & 1M* & 1000* & 1012 & FLORES-200\\ 
& Ora & Oriya & Oriya & Indo-European & Indo-Aryan & Low  & 1M* & 1000* & 1012 & FLORES-200\\ 
\hline 
\end{tabular}}
\caption{Dataset details and Statistics. * are obtained from \citet{ramesh-etal-2022-samanantar}}
\label{tab:data}
\end{table*}

\section{Model Training Details}
\label{app:impl}
We used the FairSeq library \cite{ott2019fairseq} to train proposed \textsc{CharSpan} and other baseline models. Training and implementation details are presented in Table \ref{tab:train_del}. The best checkpoint was selected based on validation loss. The training time for the Indo-Aryan family of languages was approximately 8 hours; for the Romance languages, it was approximately 7 hours, and for the Malay-Polynesian languages, it was less than 1 hour. Each language inference was completed within a time frame of less than 5 minutes. Due to computational limitations, the performance of the model was reported based on a single run. During the generation process, a batch size of 64 and a beam size of 5 were used, with the remaining parameters set to the default values provided by FairSeq. For data-pre-processing and script conversion for Indic languages, we use the Indic NLP  library\footnote{\url{https://github.com/anoopkunchukuttan/indic_nlp_library}}.

\begin{table}[!htb]
    \centering
    \scalebox{0.780}{
    \begin{tabular}{l|c}
    \hline
       architecture   & encoder-decoder (transformers) \\
       \# encoder layers  &  6 \\ 
       \# decoder layers & 6 \\
       \# parameters & 46,956,544 shared \\
        learning rate (lr) &  $5e^{-4}$ \\
      optimizer & adam \\
       dropout rate &  0.2 \\
      input size &  210 tokens (both side) \\
      epochs & 15 \\
      tokens per batch  &  32768\\
     clip-norm  & 1.0 \\
     lr scheduler &  inverse sqrt\\
     \# GPUs & 8 \\
     type of GPU & V100 Nvidia \\
     generation batch size & 64 \\
     beam size & 5 \\
    \hline
    \end{tabular}}
    \caption{Model implementation and training details }
    \label{tab:train_del}
\end{table}

\section{Performance Evaluation with BLEU, BLEURT and COMET Metrics}
\label{sec:othermetirc}
BLEU\footnote{computed with SacreBLEU BLEU signature: \textit{nrefs:1|case:mixed|eff:no|tok:13a|smooth:exp|version:2.3.1}}, BLEURT and COMET scores are reported in Table \ref{tab:resultsbleu}, \ref {tab:resultsbleurt} and \ref{tab:resultscomet}, respectively. We observe the same trends as reported in the main paper for chrF\footnote{computed with SacreBLEU chrF signature:
\textit{nrefs:1|case:mixed|eff:yes|nc:6|nw:0|space:no|version:2.3.1}}. 

\begin{table*}
\centering
\scriptsize
\scalebox{0.91}{
\begin{tabular}{|l| cccccccc | cc |cc | c|}
    \hline
\multirow{2}{*}{\textbf{Models}} 
        & \multicolumn{8}{c|}{\textbf{ Indo-Aryan}} & \multicolumn{2}{c|}{\textbf{Romance}} & \multicolumn{2}{c|}{\textbf{Malay-Polynesian}} & \multirow{2}{*}{\textbf{Average}}  \\ 
        \cline{2-13}
        & \textbf{Gom}  & \textbf{Bho}  & \textbf{Hne}  & \textbf{San}  & \textbf{Npi}  & \textbf{Mai}  & \textbf{Mag} &  \textbf{Awa} & \textbf{Cat} & \textbf{Glg} & \textbf{Jav} & \textbf{Sun} &          \\
    \hline 
    BPE & 4.36 & 10.62 & 15.76 & 3.43 & 4.36 & 9.36 & 16.7 & 15.6 & 5.23 & 22.99 & 5.74 & 6.02 & 10.01     \\
    \hline
    WordDropout & 4.62 &   11.21  & 15.71 &  4.11  &  5.47   & 9.96  &  16.76  & 16.31 &  6.19   & 22.26 &  5.90 & 6.02 &   10.37 \\
    SubwordDropout & 4.57  &  9.99  &  14.47 &  3.93 &   5.25 &   9.08 &   15.53  & 16.03 &  5.85 &   20.72  & 4.78 &   4.93 & 09.59     \\
    WordSwitchOut & 4.03 & 10.75 & 15.86 & 3.56 & 4.92 & 9.91 & 16.85 & 15.54 & 5.27 & 21.97 & 5.95 & 6.35 & 10.08
 \\
    SubwordSwitchOut & 4.13 & 10.56 & 15.93 & 3.76 & 4.49 & 9.69 & 16.61 & 16.69 & 5.19 & 23.82 & 6.02 & 6.01 & 10.24
\\ \hline
    OBPE   & 4.65  & 10.62  & 16.31  & 3.63  & 4.95 & 9.18  & 16.88 & 15.69 & 5.03 & 22.91 & 5.33 & 5.81 &  10.08       \\
    SDE & 4.77 & 10.69 & 16.21 & 3.66 & 5.42 & 9.86 & 16.80 & 16.03   & 5.47 & 23.51 & 5.88 & 6.39 &   10.39   \\

    BPE-Dropout   & 5.24 & 11.33  & 15.64  & 3.71 & 4.94 & 10.00 & 16.62 & 16.63 & 5.94 & 24.07 & 5.79 & 6.65 &   10.54       \\
    \hline
    unigram char-noise & 5.21 &  12.62  & 18.29  & 3.81 & 6.55 & 11.29 & 19.47 & 18.95 & 11.82 & 24.09 & 7.35 & 6.87 &  12.19        \\
    BPE $\rightarrow$ SpanNoise $(ours)$ & 5.39 &  13.06  & 19.00  & 4.48 & 7.01 & 13.17 & 20.30 & 19.69 & 11.91 & 24.27 & 7.51 & 7.30 &  12.75        \\
    \textsc{CharSpan} $(ours)$ & 5.77 & 13.01  & 19.52  & 4.63 & 7.13 & 13.43 & 20.81 & 20.36 & 12.21 & 24.72 & 7.52 & 7.32 &  13.03  \\
    \textsc{CharSpan} + BPE-Dropout $(ours)$ & \underline{\textbf{5.81}} & \underline{\textbf{13.81}}  & \underline{\textbf{21.03}}  & \underline{\textbf{4.64}} & \underline{\textbf{8.10}} & \underline{\textbf{14.33}} &  \underline{\textbf{22.11}} & \underline{\textbf{21.25}} & \underline{\textbf{12.64}} & \underline{\textbf{ 25.35}} & \underline{\textbf{ 7.52}} & \underline{\textbf{ 7.31}} & \underline{\textbf{13.65 }}    \\
\hline
\end{tabular}}
\caption{\small Zero-shot BLEU scores results for ELRLs $\rightarrow$ English machine translation}
\label{tab:resultsbleu}
\end{table*}

\begin{table*}
\centering
\scriptsize
\scalebox{0.9}{
\begin{tabular}{|l| cccccccc | cc |cc | c|}
    \hline
\multirow{2}{*}{\textbf{Models}} 
        & \multicolumn{8}{c|}{\textbf{ Indo-Aryan}} & \multicolumn{2}{c|}{\textbf{Romance}} & \multicolumn{2}{c|}{\textbf{Malay-Polynesian}} & \multirow{2}{*}{\textbf{Average}}  \\
        \cline{2-13}
        & \textbf{Gom}  & \textbf{Bho}  & \textbf{Hne}  & \textbf{San}  & \textbf{Npi}  & \textbf{Mai}  & \textbf{Mag} &  \textbf{Awa} & \textbf{Cat} & \textbf{Glg} & \textbf{Jav} & \textbf{Sun} &          \\
    \hline
    BPE & 0.461 & 0.494 & 0.522 & 0.414 & 0.461 & 0.494 & 0.537 & 0.549 & 0.357 & 0.495 & 0.403 & 0.401 & 0.474
  \\ \hline
    WordDropout &  0.467 & 0.502 & 0.527 & 0.419 & 0.465 & 0.497 & 0.542 & 0.565 & 0.344 & 0.496 & 0.392 & 0.391 & 0.475
 \\
    SubwordDropout & 0.454 & 0.493 & 0.513 & 0.393 & 0.459 & 0.481 & 0.526 & 0.554 & 0.319 & 0.468 & 0.382 & 0.383 & 0.460
   \\
    WordSwitchOut & 0.456 & 0.501 & 0.528 & 0.395 & 0.445 & 0.497 & 0.552 & 0.551 & 0.309 & 0.477 & 0.381 & 0.381 & 0.464
  \\
    SubwordSwitchOut & 0.459 & 0.494 & 0.519 & 0.415 & 0.455 & 0.496 & 0.535 & 0.555 & 0.365 & 0.496 &0.383 & 0.385 & 0.467  \\ \hline
    OBPE   & 0.466 & 0.496 & 0.518 & 0.419 & 0.459 & 0.491 & 0.537 & 0.551 & 0.431 & 0.428 & 0.396 & 0.381 & 0.464
 \\
    SDE & 0.486 & 0.499 & 0.515 & 0.511 & 0.496 & 0.542 & 0.543 &      0.553 & 0.440 & 0.481 & 0.406 & 0.405 &  0.489    \\
    BPE-Dropout  &  0.474 & 0.494 & 0.501 & 0.413 & 0.461 & 0.481 & 0.522 & 0.555 & 0.443 & 0.443 & 0.407 & 0.412 & 0.467
\\ \hline
    unigram char-noise &  0.471 & 0.523 & 0.547 & 0.403 & 0.456 & 0.486 & 0.571 & 0.592 & 0.495 & 0.501 & 0.403 & 0.405 & 0.487
 \\
    BPE $\rightarrow$ SpanNoise $(ours)$ &   0.469 & 0.528 & 0.553 & 0.400 & 0.459 & 0.491 & 0.579 & 0.595 & 0.499 & 0.511 & 0.405 & 0.413 & 0.491
 \\
   \textsc{CharSpan} $(ours)$ &  0.471 & 0.541 & 0.571 & 0.403 & 0.471 & 0.534 & 0.593 & 0.616 & 0.502 & 0.555 & \underline{\textbf{0.419}} & 0.422 & 0.508
 \\
     \textsc{CharSpan} + BPE-Dropout $(ours)$ & \underline{\textbf{0.478}} & \underline{\textbf{0.548}} & \underline{\textbf{0.582}} & \underline{\textbf{0.421}} & \underline{\textbf{0.478}} & \underline{\textbf{0.535}} & \underline{\textbf{0.604}} & \underline{\textbf{0.623}} & \underline{\textbf{0.505}} & \underline{\textbf{0.567}} & \underline{\textbf{0.419}} & \underline{\textbf{0.429}} & \underline{\textbf{0.515}}
 \\ 
\hline
\end{tabular}}
\caption{\small Zero-shot BLEURT (computed with \textit{BLEURT-20 } checkpoint) scores results for ELRLs $\rightarrow$ English}
\label{tab:resultsbleurt}
\end{table*}

\begin{table*}
\centering
\scriptsize
\scalebox{0.9}{
\begin{tabular}{|l| cccccccc | cc |cc | c|}
    \hline
\multirow{2}{*}{\textbf{Models}} 
        & \multicolumn{8}{c|}{\textbf{ Indo-Aryan}} & \multicolumn{2}{c|}{\textbf{Romance}} & \multicolumn{2}{c|}{\textbf{Malay-Polynesian}} & \multirow{2}{*}{\textbf{Average}}  \\
        \cline{2-13}
        & \textbf{Gom}  & \textbf{Bho}  & \textbf{Hne}  & \textbf{San}  & \textbf{Npi}  & \textbf{Mai}  & \textbf{Mag} &  \textbf{Awa} & \textbf{Cat} & \textbf{Glg} & \textbf{Jav} & \textbf{Sun} &          \\
        \hline
    BPE & 0.536 & 0.632 & 0.671 & 0.511 & 0.525 & 0.593 & 0.694 & 0.716 & 0.494 & 0.714 & 0.444 & 0.441 & 0.580
 \\    \hline
    WordDropout & 0.551 & 0.648 & 0.678 & 0.521 & 0.557 & 0.618 & 0.695 & 0.728 & 0.565 & 0.715 & 0.451 & 0.443 & 0.597
 \\
    SubwordDropout &  0.541 & 0.638 & 0.659 & 0.528 & 0.548 & 0.607 & 0.684 & 0.717 & 0.524 & 0.686 & 0.437 & 0.428 & 0.583
   \\
    WordSwitchOut & 0.544 & 0.647 & 0.681 & 0.522 & 0.563 & 0.621 & 0.706 & 0.719 & 0.529 & 0.702 & 0.453 & 0.452 & 0.594
  \\
    SubwordSwitchOut &  0.542 & 0.641 & 0.668 & 0.521 & 0.528 & 0.601 & 0.694 & 0.721 & 0.567 & 0.718 & 0.452 & 0.451 & 0.592  \\     \hline
    OBPE   &  0.541  & 0.629 & 0.667 & 0.504 & 0.527 & 0.589 & 0.691 & 0.715 & 0.492 & 0.721 &  0.363 & 0.611 & 0.587    \\
    SDE & 0.549 & 0.636 & 0.666 & 0.513 & 0.529 & 0.591 & 0.697 & 0.735    & 0.513 & 0.731 & 0.357 & 0.618 & 0.594     \\
    BPE-Dropout  & 0.549  & 0.638  & 0.644 & 0.506 & 0.531 & 0.589 & 0.677 & 0.721 & 0.504 & 0.747 & 0.373 & 0.626 & 0.592   \\
        \hline
    unigram char-noise & 0.562 &  0.679 & 0.701 & 0.536 & 0.573 & 0.634 & 0.728 & 0.754 & 0.554 & 0.741 & 0.408 & 0.621 & 0.624  \\
    BPE $\rightarrow$ SpanNoise $(ours)$ &  0.557 & 0.676 & 0.706 & 0.542 & 0.581 & 0.651 & 0.724 & 0.755 & 0.561 & 0.751 & 0.403 & 0.622 & 0.627
  \\
    \textsc{CharSpan} $(ours)$ & 0.571  & 0.695 & 0.723 & \underline{\textbf{0.556}} & 0.611 & 0.685 & 0.747 & 0.772 & 0.568 & \underline{\textbf{0.759}} & \underline{\textbf{0.417}} & 0.627 & 0.644
    \\
     \textsc{CharSpan} + BPE-Dropout $(ours)$ & \underline{\textbf{0.579}} & \underline{\textbf{0.705}} & \underline{\textbf{0.733}} & 0.551 & \underline{\textbf{0.616}} & \underline{\textbf{0.687}} & \underline{\textbf{0.757}} & \underline{\textbf{0.778}} & \underline{\textbf{0.572}} & 0.756 & 0.414 & \underline{\textbf{0.631}} & \underline{\textbf{0.648}}  \\ 
\hline
\end{tabular}}
\caption{\small Zero-shot COMET (computed with \textit{Unbabel/wmt22-comet-da} model) scores results for ELRLs $\rightarrow$ English}
\label{tab:resultscomet}
\end{table*}


\begin{table*}
\centering
\scriptsize
\scalebox{0.97}{
\begin{tabular}{l| cccc | cccc | cccc}
    \toprule
        XX $\rightarrow$ EN & \multicolumn{4}{c|}{\textbf{Indo-Aryan}} & \multicolumn{4}{c|}{\textbf{Romance}} & \multicolumn{4}{c}{\textbf{Malay-Polynesian}}   \\     \midrule 
\multirow{2}{*}{\textbf{Models}} 
        & \multicolumn{2}{c}{\textbf{BLEU}} & \multicolumn{2}{c|}{\textbf{chrF}} & \multicolumn{2}{c}{\textbf{BLEU}} & \multicolumn{2}{c|}{\textbf{chrF}} & \multicolumn{2}{c}{\textbf{BLEU}} & \multicolumn{2}{c}{\textbf{chrF}} \\
        \cline{2-13}
        & \textbf{Hin}  & \textbf{Mar}  & \textbf{Hin}  & \textbf{Mar} & \textbf{Spa}  & \textbf{Pot}  & \textbf{Spa}  & \textbf{Pot} & \textbf{Ind} & \textbf{Zsm} & \textbf{Ind} & \textbf{Zsm}         \\
    \midrule 
    BPE &    37.44 & 26.31 & 64.04 & 54.47 & 41.44 & 35.38 & 68.71 & 63.27 & 29.61 & 21.76 & 58.31 & 49.14
    \\\midrule
    WordDropout  & 36.54 & 26.31 & 63.27 & 53.96 & 39.32 & 32.73 & 66.89 & 60.86 & 27.59 & 20.42 & 56.72 & 48.22
 \\
    SubwordDropout   &  36.64 & 26.22 & 63.46 & 54.57 & 39.84 & 33.04 & 67.56 & 61.58 & 26.73 & 18.80 & 57.02 & 48.82
 \\
    WordSwitchOut &  34.12 & 23.84 & 60.98 & 51.84 & 35.27 & 30.63 & 63.25 & 58.38 & 27.04 & 19.60 & 55.69 & 46.93
\\
    SubwordSwitchOut  &37.11 & 26.03 & 63.78 & 54.06 & 42.26 & 35.68 & 68.65 & 62.97 & 27.12 & 19.76 & 55.72 & 47.34 \\
    \midrule
    OBPE  & 37.32 & 26.90  & 64.05  & 55.03 & 41.81 & 36.44 & 68.17 & 63.45 & 28.14 & 21.83 & 57.11 & 49.21        \\
    SDE &  37.22 & 26.19  & 63.98 & 55.44 & 41.41 & 35.51 & 68.61 & 62.89 &  29.11 & 21.52  & 58.25 & 48.98\\
    BPE-Dropout   & 37.22 & 26.93  &  64.11 & 55.31 &  41.88 & 36.72  & 68.06 & 63.79 &  30.39 & 22.54  & 59.33  & 50.17  \\
    \midrule
    unigram char-noise & 37.05 &  26.95 &  63.81 & 54.83 &  39.83 & 32.91  & 67.62 & 61.24 &  28.79 & 22.01  & 57.65 & 49.91  \\
    BPE  $\rightarrow$ SpanNoise $(ours)$ &  36.66 & 26.93  & 63.80 & 54.84 &  39.92 &  32.22 & 66.83 & 61.06 & 27.84 & 22.16   & 57.15 & 50.19 \\
   \textsc{CharSpan} $(ours)$ & 36.68 & 26.70  &  63.87 & 54.59 &  40.04 & 32.36  & 66.95 &  61.03 &  27.84 & 21.87  & 56.75 & 49.58 \\
   \textsc{CharSpan} + BPE-Dropout $(ours)$ & 37.62 & 27.10  & 64.15  & 55.03 & 41.21 & 33.64 & 66.90 & 61.39 &  28.91 & 22.26  & 57.99 & 50.59 \\
\bottomrule
\end{tabular}}
\caption{BLEU and chrF Scores: High resource language performance for all three language families. It can be observed that, even with the inclusion of noise augmentation, the proposed model exhibits only a slight decrease in performance for HRLs.}
\label{tab:hrlresults}
\end{table*}

\begin{table*}
\centering
\scriptsize
\scalebox{0.97}{
\begin{tabular}{l| cccc | cccc | cccc}
    \toprule
        XX $\rightarrow$ EN & \multicolumn{4}{c|}{\textbf{Indo-Aryan}} & \multicolumn{4}{c|}{\textbf{Romance}} & \multicolumn{4}{c}{\textbf{Malay-Polynesian}}   \\     \midrule 
\multirow{2}{*}{\textbf{Models}} 
        & \multicolumn{2}{c}{\textbf{BLEURT}} & \multicolumn{2}{c|}{\textbf{COMET}} & \multicolumn{2}{c}{\textbf{BLEURT}} & \multicolumn{2}{c|}{\textbf{COMET}} & \multicolumn{2}{c}{\textbf{BLEURT}} & \multicolumn{2}{c}{\textbf{COMET}} \\
        & \textbf{Hin}  & \textbf{Mar}  & \textbf{Hin}  & \textbf{Mar} & \textbf{Spa}  & \textbf{Pot}  & \textbf{Spa}  & \textbf{Pot} & \textbf{Ind} & \textbf{Zsm} & \textbf{Ind} & \textbf{Zsm}         \\
    \midrule 
    BPE & 0.775 & 0.726 & 0.891 & 0.857 & 0.769 & 0.720 & 0.871 & 0.830 & 0.687 & 0.561 & 0.821 & 0.701
 \\\midrule
    WordDropout  & 0.774 & 0.725 & 0.891 & 0.854 & 0.755 & 0.701 & 0.86 & 0.814 & 0.681 & 0.555 & 0.815 & 0.693
 \\
    SubwordDropout   &  0.773 & 0.725 & 0.889 & 0.854 & 0.757 & 0.691 & 0.861 & 0.806 & 0.672 & 0.548 & 0.803 & 0.683
  \\
    WordSwitchOut &  0.756 & 0.706 & 0.879 & 0.842 & 0.707 & 0.651 & 0.826 & 0.775 & 0.665 & 0.547 & 0.804 & 0.688
 \\
    SubwordSwitchOut  &  0.776 & 0.724 & 0.892 & 0.855 & 0.771 & 0.721 & 0.872 & 0.833 & 0.663 & 0.548 & 0.801 & 0.687
 \\\midrule
     OBPE  & 0.777 & 0.731 & 0.893 & 0.861 & 0.766 & 0.727 & 0.863 & 0.821 & 0.672 & 0.551 & 0.811 & 0.697
  \\
    SDE & 0.772 & 0.721  & 0.889 & 0.856 & 0.765 & 0.721 & 0.866 & 0.832 & 0.679  & 0.558  & 0.818 & 0.699\\
    BPE-Dropout  &  0.773 & 0.727 & 0.891 & 0.858 & 0.772 & 0.7281 & 0.881 & 0.839 & 0.706 & 0.586 & 0.838 & 0.729
 \\\midrule
    unigram char-noise & 0.775 & 0.731 & 0.892 & 0.857 & 0.756 & 0.683 & 0.861 & 0.798 & 0.681 & 0.574 & 0.815 & 0.716
 \\
    BPE $\rightarrow$ SpanNoise $(ours)$ & 0.773 & 0.728 & 0.891 & 0.857 & 0.755 & 0.685 & 0.861 & 0.801 & 0.685 & 0.581 & 0.821 & 0.724
 \\
    \textsc{CharSpan} $(ours)$  & 0.775 & 0.726 & 0.892 & 0.856 & 0.755 & 0.681 & 0.861 & 0.799 & 0.671 & 0.569 & 0.829 & 0.714
 \\
    \textsc{CharSpan} + BPE-Dropout $(ours)$ & 0.775 & 0.726 & 0.892 & 0.856 & 0.768 & 0.683 & 0.877 & 0.801 & 0.685 & 0.582 & 0.823 & 0.726 \\
\bottomrule
\end{tabular}}
\caption{BLEURT and COMET Scores: High resource language performance for all three language families}
\label{tab:hrlresults2}
\end{table*}

\begin{table*}
\centering
\scriptsize
\scalebox{1.1}{
\begin{tabular}{l| ccccccc | c}
    \toprule
\multirow{2}{*}{\textbf{Experimental Setup}} 
        & \multicolumn{7}{c|}{\textbf{ Indo-Aryan}} & \textbf{Average}  \\ 
        & \textbf{Bho}  & \textbf{Hne}  & \textbf{San}  & \textbf{Npi}  & \textbf{Mai}  & \textbf{Mag} &  \textbf{Awa} &          \\
    \midrule 
    \multicolumn{9}{c}{\textbf{ChrF Scores}} \\
    \hline
    \textsc{CharSpan} with Hin, Mar, Pan, Guj, Ben & 38.81 & 45.39 & 30.34 & 34.4 &  41.67 & 45.82 & 43.78 & 40.03  \\ 
    \textsc{CharSpan} with Hin, Mar, Pan, Guj & 37.68 & 43.49 & 28.44 & 32.22 & 39.43 & 44.34 & 42.33 & 38.27   \\ 
   \textsc{CharSpan} with Hin, Mar, Pan& 33.32 & 38.81 & 25.71 & 29.21 & 54.82 & 39.17 & 26.47 & 35.35  \\ 
    \textsc{CharSpan} with Hin, Mar & 29.70 &  33.13 & 23.83 & 26.12 & 31.88 & 33.83 & 33.13 & 30.23   \\ 
    \textsc{CharSpan} with Hin & 20.96 &  21.92 & 15.90 &  17.97 & 20.85 & 22.85 & 21.75 & 20.31   \\ 
    \hline
    \multicolumn{9}{c}{\textbf{BLEU Scores}} \\
    \hline
   \textsc{CharSpan} with Hin, Mar, Pan, Guj, Ben & 10.46 & 15.97 & 4.87 &  7.02 &  11.83 & 16.32 & 14.65 & 11.58  \\ 
    \textsc{CharSpan} with Hin, Mar, Pan, Guj & 9.55 &  14.32 & 3.92 &  5.99 &  9.85 &  14.71 & 13.47 & 10.25  \\ 
    \textsc{CharSpan} with Hin, Mar, Pan& 7.41 &  10.21 & 2.91 &  4.63 &  7.88 &  11.01 & 9.89 &  7.70
  \\ 
   \textsc{CharSpan} with Hin, Mar & 5.30 & 7.06 &  2.40 & 3.20 &  5.00 & 7.28 &  6.96 &  5.31    \\ 
    \textsc{CharSpan} with Hin & 2.03 &  2.27 &  0.6 & 0.97 &  1.77 &  2.23 &  2.39 &  1.75  \\ 
\bottomrule
\end{tabular}}
\caption{\small Zero-shot multilingual performance of char-span noise augmentation model. We have considered multiple combinations of high-resource languages for a multilingual setup. Due to computational constraints, 1 million parallel training data for each language was considered. All the languages are considered from the FLORES-200 test set.}
\label{tab:resultmultilingual}
\end{table*}


\begin{figure*}
    \centering
     \scriptsize
    \includegraphics[scale=0.7]{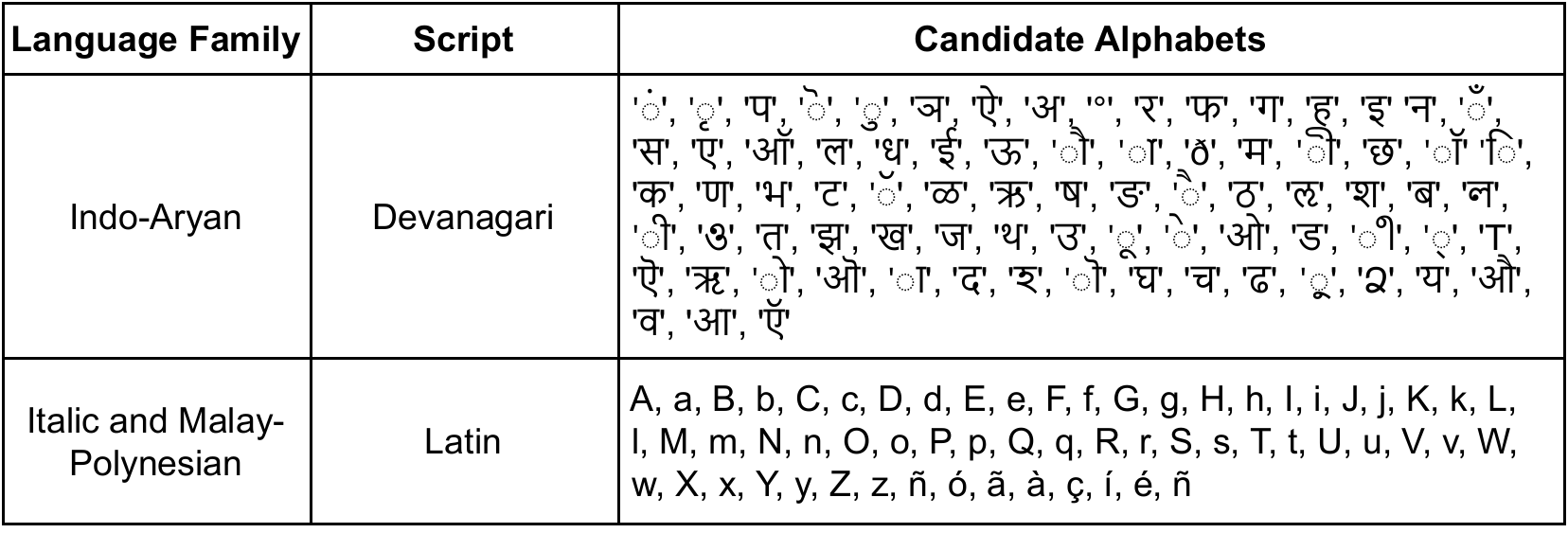}
    \caption{Candidate alphabets for noise augmentation. For the Indo-Aryan language family, the Devanagari alphabet is used, while the Latin alphabet is employed for the Romance and Malay-Polynesian language families.}
    \label{tab:candalpha}
\end{figure*}




\begin{table*}
\scriptsize
\centering
\begin{tabular}{l| ccc | ccc}
    \toprule
\multirow{2}{*}{\textbf{Experimental Setups}} 
        & \multicolumn{3}{c|}{\textbf{BLEU (XX $\rightarrow$ EN)}} & \multicolumn{3}{c}{\textbf{chrF (XX $\rightarrow$ EN)}} \\
        & \textbf{Gom}  & \textbf{Bho}  & \textbf{Hne}  & \textbf{Gom}  & \textbf{Bho}  & \textbf{Hne}         \\
    \midrule 
    char-noise ($9$\%-$11$\% + replacement with only vowels)   & 4.77 & 11.21 & 15.17  & 28.08 & 40.36 & 46.13        \\
    char-noise ($9$\%-$11$\%+ replacement with only consonants) &  4.79 & 11.25 & 15.3 & 26.95 & 40.51 &  46.17       \\
    char-noise ($9$\%-$11$\% + replacement with char sound similarity ) & 4.55 & 10.7 & 15.78  & 27.86 & 40.45 &  46.98    \\
    char-noise ($9$\%-$11$\% + with number and punctuation)  & 5.13 & 12.07 & 17.66 & 27.66 & 41.43 & 48.68  \\
    \midrule
    char-noise ($9$\%-$11$\% + only insertion)  &  5.04 & 12.3 & 17.81  & 27.50 & 41.87 & 48.74  \\
    char-noise ($9$\%-$11$\% + only replacement)  &  5.58 & 12.8 & 18.75 & 28.85 & 42.43 & 49.68  \\
    char-noise ($9$\%-$11$\%+ only deletion)  &  4.22 &11.92 & 18.39 & 28.65 & 42.02 & 49.36 \\
    \midrule
    char-noise ($4$\%-$6$\% + all three operations + equal probability) & 5.44 & 11.66 & 18.01  & 28.62 & 40.95 &  48.63     \\
    char-noise ($14$\%-$16$\% + all three operations + equal probability)  &  5.17 & 11.4 & 17.01  & 27.93 & 40.32 &  47.61      \\
    char-noise ($9$\%-$11$\% + all three operations + equal probability) &  5.21 & 12.62 & 18.29 & 28.85 & 42.53 & 49.35\\
    \midrule
    char-span noise ($9$\%-$11$\% + 1-3 grams + replacement: N random chars -> span )  & 3.80 & 8.80 & 13.11 & 25.38 & 28.22 & 43.39  \\
    char-span noise ($9$\%-$11$\% + 1-3 grams + insertion: 1 random chars -> span )  &  \underline{\textbf{5.84}} &13.29 & 20.49 & 29.29 & 43.51 & 51.33 \\
    char-span noise ($9$\%-$11$\% + 1-3 grams + insertion: N random chars -> span )  &  4.81 & 12.21 & 17.36  & 26.98 & 41.26 &  47.91 \\
    char-span noise ($9$\%-$11$\% + 1-3 grams + all three operations + equal probability)  & 4.01 & 10.41 & 16.33  & 27.99 & 36.66 & 46.13  \\
    \midrule
    char-span noise ($9$\%-$11$\% + 1-2 grams +  replacement and deletion + equal probability)  &  5.42 & 12.08 & 18.02  & 29.17 & 42.21 & 49.17  \\
    char-span noise ($9$\%-$11$\% + 1-4 grams +  replacement and deletion + equal probability)  &  5.79 & 11.85 & 18.02 & \underline{\textbf{29.71}} & 42.41 & 49.74  \\
    char-span noise ($9$\%-$11$\% + 1-5 grams +  replacement and deletion + equal probability)  &  5.56 & 11.36 & 17.06 & 24.13 & 26.35 &  29.55 \\
    char-span noise ($9$\%-$11$\%+  1-3 grams +  replacement and deletion +unequal probability )  & 5.48 & 12.12 & 18.16 & 29.01 & 41.74 & 49.37  \\
    
    \midrule 
    \textbf{Proposed:} char-span noise ( $9$\%-$11$\% + 1-3 grams +  replacement and deletion + equal probability)  & \underline{\textbf{5.81}} & \underline{\textbf{13.81}}  & \underline{\textbf{21.03}}  & \underline{\textbf{29.71}}  & \underline{\textbf{43.75}} & \underline{\textbf{51.69}}  \\
\bottomrule
\end{tabular}
\caption{Ablation Study and Different Experimental Setups. Similar trends were observed for other ELRLs and language families. Approximately 200 experiments were performed.}
\label{tab:ablation}
\end{table*}

\section{Language Similarity Histogram}
\label{app:langsim}
\begin{figure*}
    \centering
    \includegraphics[width=0.32\textwidth]{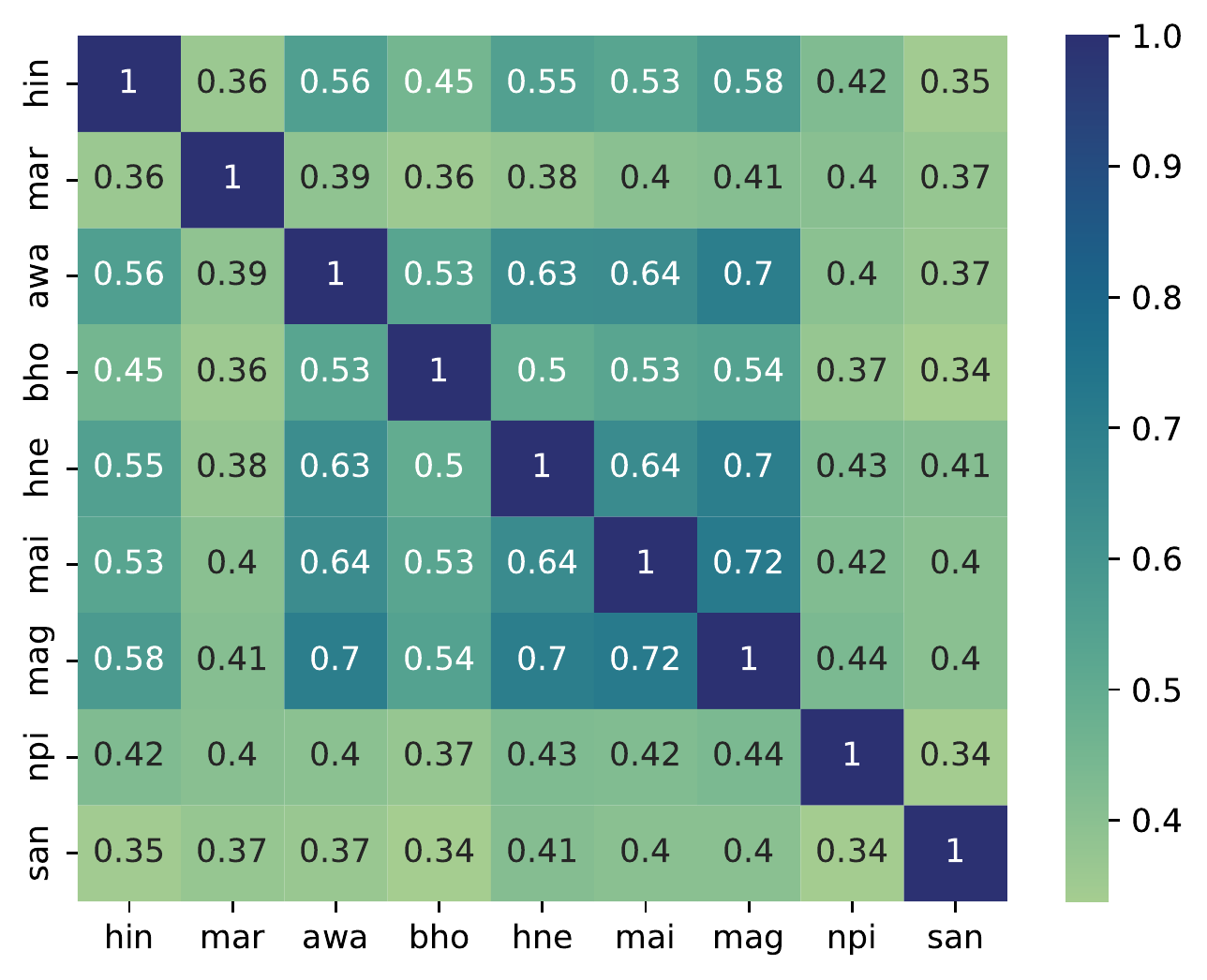}
    \includegraphics[width=0.32\textwidth]{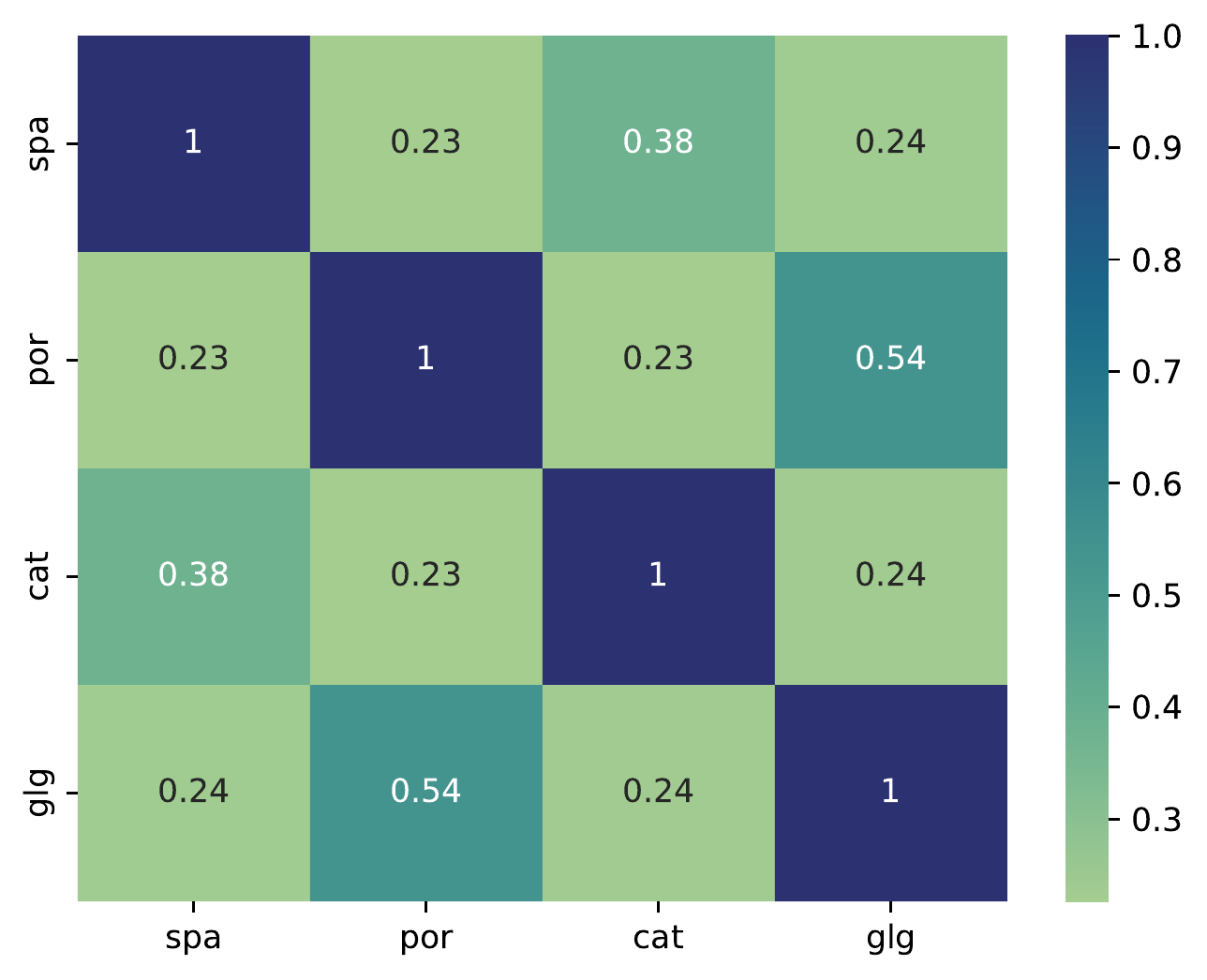}
    \includegraphics[width=0.32\textwidth]{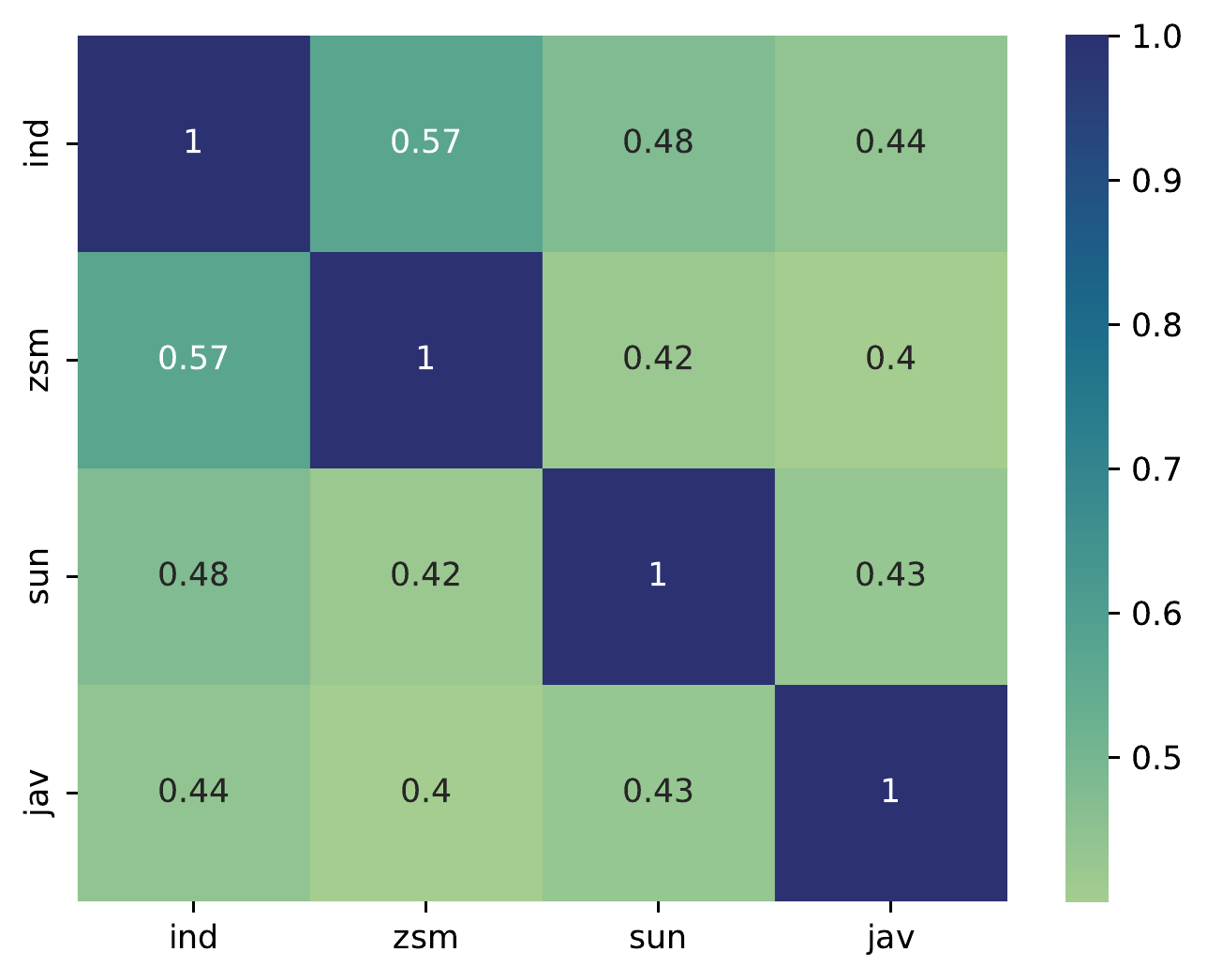}
    \caption{Lexical similarity (LCSR) heatmaps for three languages families. The Indo-Aryan languages are considered to use the Devanagari script, while the Latin script is used by the other two language families.}
    \label{fig:langsim}
\end{figure*}

\begin{figure*}
    \captionsetup{font=scriptsize}
    \centering
     \scriptsize
    \includegraphics[scale=0.6]{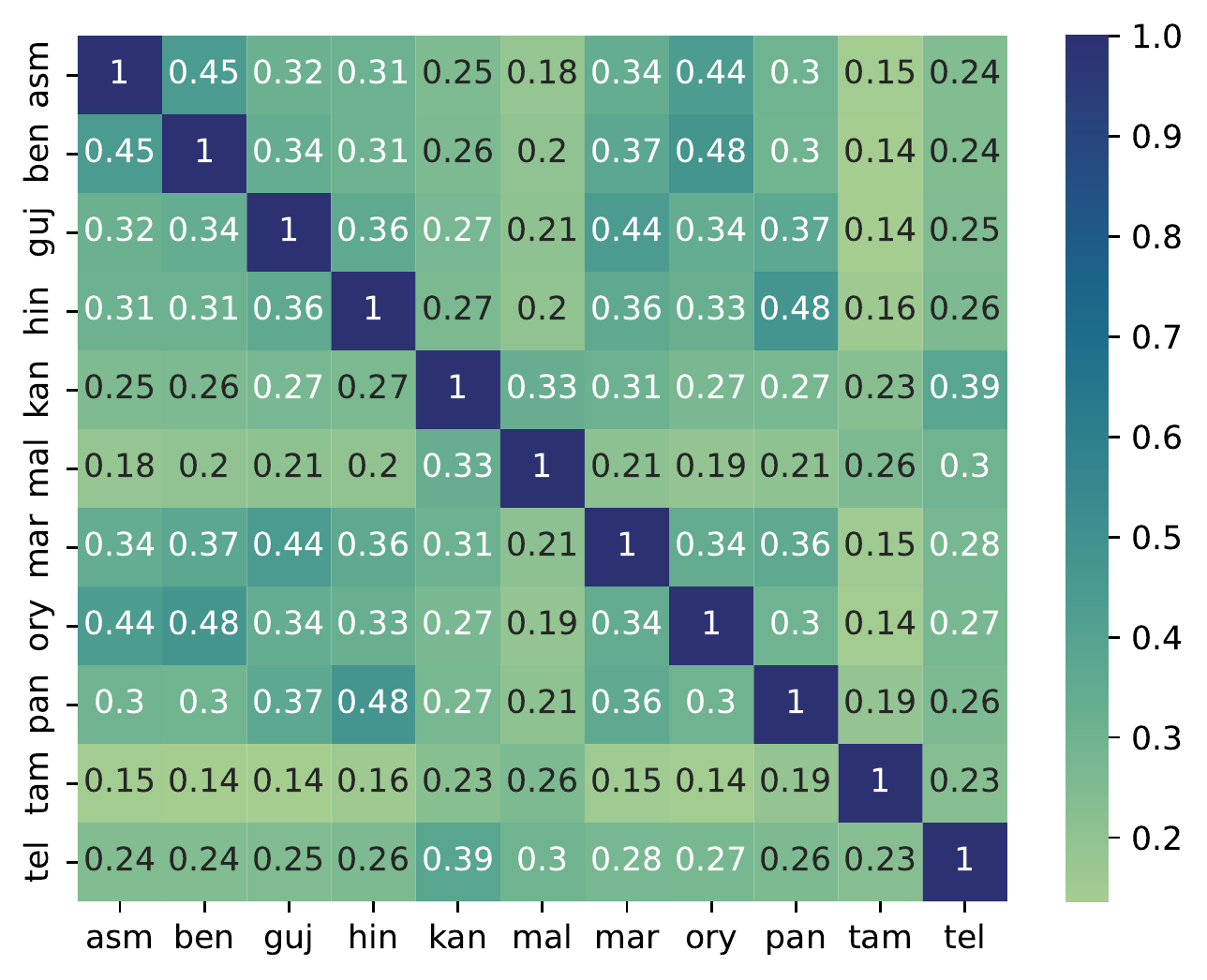}
    \caption{Lexical similarity heatmap for additional languages used in the analysis section. Here we have shown similarity scores for Assamese (asm), Bengali (ben), Gujrati (guj), Panjabi (pan), Hindi (him), Marathi (mar), Oriya (ory), Malayalam (mal), Kannada (kan), Tamil (tam) and Telugu (tel) languages.}
    \label{fig:lexicalsimilarity_others}
\end{figure*}

As depicted in Fig. \ref{fig:langsim}, a similarity analysis in the form of a heatmap for the selected language families and languages is presented. The analysis shows that extremely low-resource languages (ELRLs) are closely related to high-resource languages (HRLs). The lexical similarity between languages was measured using character-level longest common subsequence ratio (LCSR) metric \cite{melamed-1995-automatic}. The similar heat map is also presented for less similar languages in Fig. \ref{fig:lexicalsimilarity_others}. These languages were used in the multiple analyses.

\begin{figure*}
    \centering
    \includegraphics[scale=0.6]{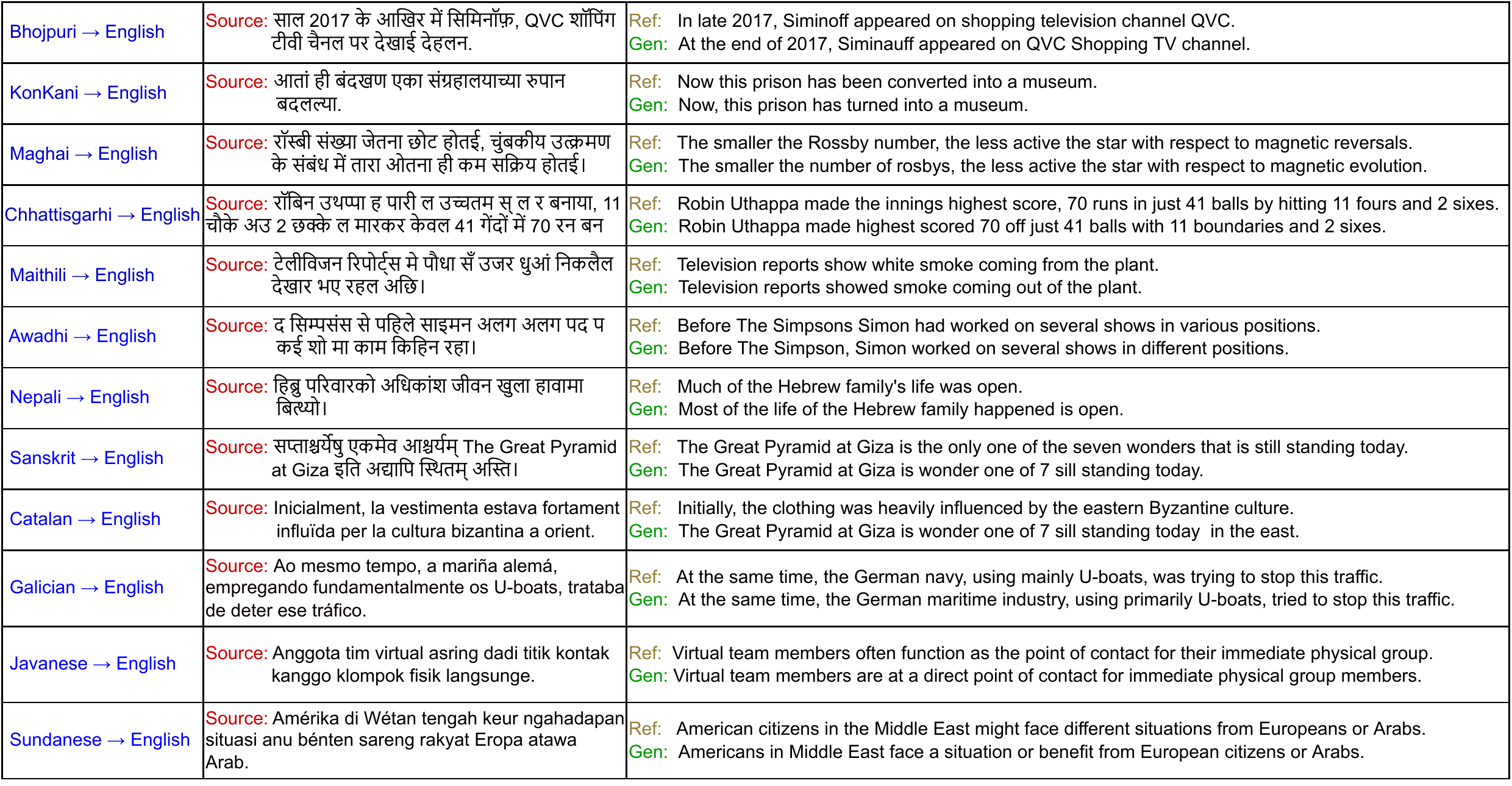}
    \caption{Zero-shot Sample generations with \textsc{CharSpan} model for ELRLs.}
    \label{fig:sample}
\end{figure*}

\section{Impact of Additional Small ELRLs parallel Data}
\label{sec:smallpd}
Here, we combined small ELRL parallel data with the HRLs training data for BPE and \textsc{CharSpan} model. The results are presented in Table \ref{tab:addparelrls}. The inclusion of additional data boosts both model performance, and \textsc{CharSpan} still outperforms the BPE model. 

\begin{table*}
\centering
\begin{tabular}{|l|c|c| c| c| c| c|}
    \hline
        \textbf{Setup} & \textbf{Gom} & \textbf{Bho}  & \textbf{Hne}   & \textbf{San} & \textbf{Npi}  & \textbf{Mai} \\
    \hline
    BPE & 26.75 & 39.75 & 46.57 & 27.97 &  30.84 & 39.79    \\
    BPE+ELRL$_\textrm{par}$ & 26.54 & 42.66 &  52.52 & 31.88  &  38.09  & 43.22   \\
    \textsc{CharSpan} & 29.71 & 43.75 & 51.69 & 31.40 &  36.52 & 45.84     \\
    \textsc{CharSpan}+ELRL$_\textrm{par}$ & 29.65 & 45.39 & 53.38 &  33.92 & 39.66 & 47.18 \\
    \hline
\end{tabular}
\caption{\small Translation quality (chrF) with an additional 1000  ELRL-English parallel sentences (ELRL$_\textrm{par}$).}
\label{tab:addparelrls}
\end{table*}

\section{Effect of Cross-Lingual Transfer}
\label{sec:cltfull}
We did the following studies to understand why noise helps. The effectiveness of cross-lingual transfer depends on how well-aligned the representations of the HRL and ELRL are. Our hypothesis is that regularization with \textit{char-level noise brings the representations of the HRL and ELRL closer to each other, thus improving cross-lingual transfer}. To measure these, we computed the cosine similarity of encoder representations from parallel HRL and ELRL sentences of 3 different models (baseline BPE, Unigram character-noise, \textsc{CharSpan}). The encoder representations were computed by mean-pooling the token representations of the top layer of the encoder. The Table -\ref{tab:resultschrfappndix} shows the results (we report average results over the test set). We can clearly see that the similarity of encoder representations significantly increases in noise-augmented models. Further, \textsc{CharSpan} improves over unigram char-noise, reflecting improved translation quality.

\begin{table*}[!htb]
\centering
\resizebox{0.9\textwidth}{!}{  
\begin{tabular}{|l| ccccccc | cc |cc | c|}
    \hline
\multirow{2}{*}{\textbf{Models}} 
        & \multicolumn{7}{c|}{\textbf{ Indo-Aryan}} &
        \multicolumn{2}{c|}{\textbf{Romance}} & \multicolumn{2}{c|}{\textbf{Malay-Polynesian}} & \multirow{2}{*}{\textbf{Average}}  \\
        \cline{2-12}
        & \textbf{Bho}  & \textbf{Hne}  & \textbf{San}  & \textbf{Npi}  & \textbf{Mai}  & \textbf{Mag} &  \textbf{Awa} & \textbf{Cat} & \textbf{Glg} & \textbf{Jav} & \textbf{Sun} &          \\
    \hline
    BPE & 0.761 & 0.793 & 0.701 & 0.744 & 0.762 & 0.809 & 0.792 & 0.721 & 0.813 & 0.731 & 0.736 & 0.760 \\
     UCN & 0.853 & 0.888 & 0.765 & 0.821 & 0.849 & 0.897 & 0.883 & 0.803 & 0.879 & 0.813 & 0.811 &  0.842 \\
   \textsc{CharSpan}  & \textbf{0.871} &\textbf{0.909} & \textbf{0.789} & \textbf{0.858} & \textbf{0.868} & \textbf{0.913} & \textbf{0.901} & \textbf{0.831} & \textbf{0.903} & \textbf{0.846} & \textbf{0.856} &  \textbf{0.867}\\
\hline
\end{tabular}}
\caption{Average cosine similarity between representations of source HRLs and source LRLs. UNC: Unigram Char-Noise }
\label{tab:resultschrfappndix}
\end{table*}

\section{Error Analyses}
\label{sec:errana}
\subsection{Basline Generations are Transliterated}
\begin{figure*}
    \centering
    \includegraphics[scale=0.7]{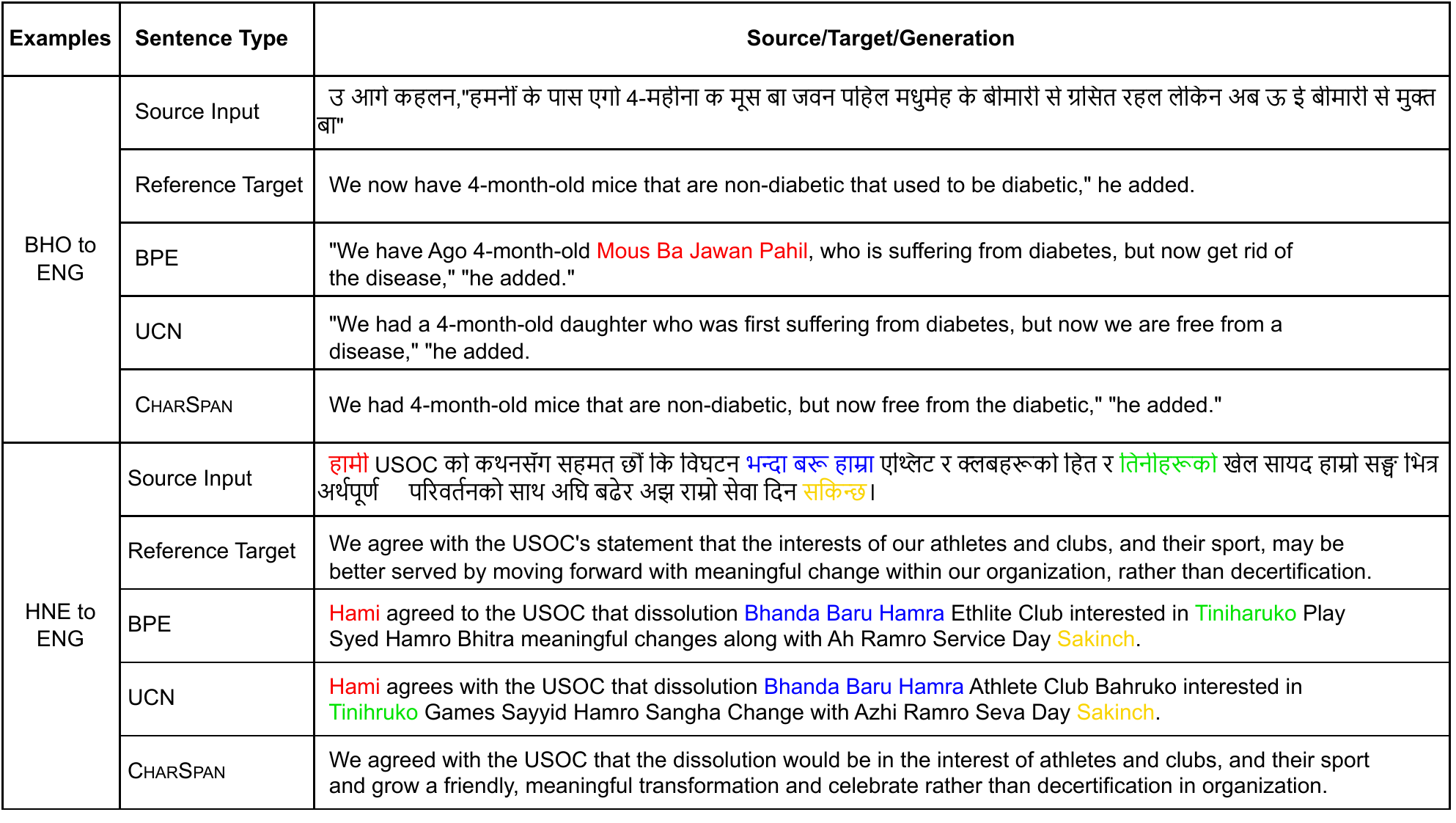}
    \caption{The generation errors (transliteration) from different baseline models. The proposed \textsc{CharSpan} model successfully mitigates those errors. Colors indicate the corresponding transliteration in a generation.}
    \label{fig:erroranalysis1}
\end{figure*}

Fig. \ref{fig:erroranalysis1} presents a few sample examples where baseline models give generation error. Here, we look for transliteration errors. It can observed that many of the source words are directly transliterated in target generation for baseline models; however, the proposed \textsc{CharSpan} model successfully mitigates these errors. 

\subsection{Grammatical Well-Formedness}
It is often observed that the generations are grammatically not sound, and such features are easily missed by performance evaluation metrics like ChrF and BLEU. With this error analysis, we aim to investigate the grammatical well-formedness of generations from different baseline models. To score the grammatical well-formedness, we use L'AMBRE tool\footnote{\url{https://github.com/adithya7/lambre}}. The results are reported in Table \ref{tab:errorana2}. For simplicity, we have shown results for only the Indo-Aryan family. The \textit{CharSpan} shows better Grammatical formation than BPE and Unigram char-noise model across all ELRL.

These error analyses further prove that the performance gains are genuine for the \textsc{CharSpan} model. 

\begin{table*}[!htb]
\centering
\resizebox{0.7\textwidth}{!}{  
\begin{tabular}{|l| ccccccc |}
    \hline
\multirow{2}{*}{\textbf{Models}} 
        & \multicolumn{7}{c|}{\textbf{ Indo-Aryan}} \\
        \cline{2-8}
        & \textbf{Bho}  & \textbf{Hne}  & \textbf{San}  & \textbf{Npi}  & \textbf{Mai}  & \textbf{Mag} &  \textbf{Awa}          \\
    \hline
    BPE & 0.9782 & 0.9813 & 0.9444 & 0.9624 & 0.9647 & 0.9784 & 0.9812 \\
     UCN & 0.9754 & 0.9616 & 0.9504 & 0.9592 & 0.947 & 0.9708 & 0.9753  \\
   \textsc{CharSpan}  & \textbf{0.9856} & \textbf{0.9865} & \textbf{0.9658} & \textbf{0.9735} & \textbf{0.9802} & \textbf{0.9842} &\textbf{ 0.9836} \\
\hline
\end{tabular}}
\caption{Grammatical Well-Formedness for different models with L'AMBRE}
\label{tab:errorana2}
\end{table*}

\vspace{4cm}
\section{Literature Review}
In this section, we presented details of three threads of literature review related to the proposed work. This is summarized in Section \ref{sec:related} of the main paper.  
\subsection{MT for Low-resource Languages}
Due to the unavailability of the large bi-text dataset for low-resource languages, much of the existing research focuses on \textit{multilingual} MT. This enables cross-lingual transfer \cite{nguyen-chiang-2017-transfer, zoph-etal-2016-transfer} and allows related languages to learn from each other \cite{fan2021beyond,costa2022no,siddhant2022towards}. While this direction has gained significant attention, the performance improvement for LRLs as compared to HRLs has been limited \cite{tran-etal-2021-facebook} and remains an open area of research.  In another thread, efforts have been made for MT models directly from the monolingual dataset \cite{artetxe2018iclr, lample2017unsupervised,lewis-etal-2020-bart}. These unsupervised approaches show promise but still require a large amount of monolingual data, which should ideally match the domain of the HRLs  \cite{marchisio-etal-2020-unsupervised}. However, for many LRLs, monolingual datasets are not available \cite{artetxe-etal-2020-cal}. In contrast, we propose a model that does not require any bi-text/monolingual dataset and is scalable to any number of LRLs/dialects.

\subsection{Vocabulary Adaptation for MT}

Early exploration of character-based MT  showed the promise \cite{chung-etal-2016-character, 10.1162/tacl_a_00067} with coverage and robustness \cite{provilkov-etal-2020-bpe, libovicky-fraser-2020-towards}. However, recent modeling concludes a number of challenges \cite{gupta2019character, libovicky-fraser-2020-towards} in terms of training/inference times and performance as compared to the subwords models. Specifically, \citet{shaham-levy-2021-neural} shows that character MT and Byte MT \cite{costa-jussa-etal-2017-byte} have worse performance than the  Byte Pair Encoding (BPE; \cite{sennrich-etal-2016-neural}) model and limits their practical usage \cite{libovicky-etal-2022-dont}. The effectiveness of the BPE algorithm \cite{10.5555/177910.177914} is reported for NMT \cite{sennrich-etal-2016-neural} and serval other NLP tasks \cite{liu2019roberta}. Other algorithms like Sentencepiece \cite{kudo-richardson-2018-sentencepiece} and Wordpiece \cite{wordpiece2018} are similar to BPE. We take inspiration from existing works and proposed a model on BPE.

Given the potential of the BPE model, various methodologies have been developed for vocabulary modification/generation/adaption \cite{provilkov-etal-2020-bpe, khemchandani-etal-2021-exploiting, patil-etal-2022-overlap, minixhofer-etal-2022-wechsel}. In particular, the work of \citet{provilkov-etal-2020-bpe} utilizes the BPE algorithm to generate the vocabulary and sample different segmentations during training. \citet{patil-etal-2022-overlap} introduce an extension of BPE, referred to as Overlapped BPE (OBPE), which takes into account both HRLs and LRLs tokens during vocabulary creation. They demonstrate the effectiveness of this approach in only  NLU tasks. In contrast, in this study, we adopt the standard BPE model on noisy HRL data for the MT task.

\subsection{Surface/Lexical Level Noise for MT} 
Several previous studies \cite{sperber2017toward, koehn-knowles-2017-six, karpukhin-etal-2019-training, vaibhav-etal-2019-improving} have examined the use of noise augmentation strategies, including substitution, deletion, insertion, flip, and swap, at various levels of text granularity for machine translation. These strategies are explored to stabilize/improve the robustness of the model with naturally occurring noises, such as spelling mistakes. Further, these noising schemes are utilized to obtain non-canonical text in adversarial settings \cite{heigold-etal-2018-robust}. Close to ours, \citet{aepli-sennrich-2022-improving} proposed a character-based noise model to transfer the supervision from HRLs to LRLs in a zero-shot setting. They evaluated the proposed model on two NLU tasks with the pre-trained model. Unlike this, we have trained the model from scratch for the machine translation task, which is very different and more challenging than NLU tasks. Moreover, we explore the \textit{span-denoise}, which outperformed char denoise-based models and emerged as a desirable MT model for extremely low-resource languages and dialects.     


\end{document}